\documentclass[journal]{IEEEtran}

\usepackage{verbatim}
\usepackage{amsmath}

\usepackage[utf8]{inputenc}
\usepackage{graphicx}
\usepackage{subfigure}
\usepackage{float}
\usepackage{bm}
\usepackage{amssymb}
\usepackage{tcolorbox}
\usepackage{xcolor}
\usepackage{threeparttable}
\usepackage{multirow}
\usepackage{booktabs}
\usepackage[table,xcdraw]{xcolor}

\usepackage{graphics} 
\usepackage{epsfig} 
\usepackage[T1]{fontenc}
\usepackage{cite}
\usepackage{tabularray}
\usepackage{tabularx}
\usepackage{algorithm}
\usepackage{algorithmic}
\usepackage{color}
\usepackage[colorlinks=true]{hyperref}

\graphicspath{{Images/}}

\title{ATP: Anatomical Torque with Passivity-based Control Framework for Safe Upper-Limb Exoskeleton Assistance
}
\author{Yu Chen, Gong Chen and Xiang Li
\thanks{Y. Chen and X. Li are with the Department of Automation, Tsinghua University, China. G. Chen is with the Shenzhen MileBot Robotics Co., Ltd, China. This work was supported in part by the Science and Technology Innovation 2030-Key Project under Grant 2021ZD0201404, in part by
the Institute for Guo Qiang, Tsinghua University, and in part by the National Natural Science Foundation of China under Grant U21A20517 and 52075290. 
Corresponding author: Xiang Li (xiangli@tsinghua.edu.cn)}%
}

\begin{document}

\maketitle
\pagestyle{plain}  
\thispagestyle{plain} 

\begin{abstract}
Providing assistance across diverse movement tasks is a central objective of exoskeletons, and incorporating anatomical knowledge can enable responsive support that generalizes across tasks. 
However, anatomical assistance has primarily been investigated for lower-limb exoskeletons, where periodic, weight-bearing movements impose lower requirements on torque precision.
Developing a unified approach for complex and non-periodic upper-limb movements therefore remains an open challenge.
This paper proposes the Anatomical Torque with Passivity-Based Control (ATP) framework for safe upper-limb exoskeleton assistance. In ATP, anatomical reference torques are generated from a musculoskeletal system, while interaction safety is ensured by preserving passivity in human–robot interaction. 
First, a scalable musculoskeletal simulation framework is developed to train a unified reinforcement-learning-based muscle controller that generalizes across various upper-limb movements and generates anatomical reference torques without complex biomechanical computations.
Second, an online torque-refinement scheme adapts the anatomical reference torque to the high diversity of upper-limb movements, while suppressing tendon-induced spikes and incorporating a learned anomaly score to ensure safe and comfortable assistance.
Third, an interaction torque controller is developed for a cable-driven compliant exoskeleton to deliver assistance without constraining upper-limb motion to predefined trajectories, while an energy tank preserves passivity and rigorous theoretical analysis guarantees closed-loop torque tracking subject to passivity.
Simulation and real-world experiments demonstrate that the muscle controller accurately tracks a dataset containing long-duration motion sequences and generalizes to real-time human movements in physical operation. The interaction-torque controller achieves accurate torque tracking while preserving system passivity and resumes the tracking objective following energy-tank replenishment. 
Moreover, an electromyography study involving five participants shows that ATP reduces the activity of the primary target muscles during both static and dynamic tasks compared with gravity-compensation and open-loop assistance, achieving a reduction of up to $48\%$ relative to movement without the exoskeleton during a dynamic multi-joint task.
The source code is available at \href{https://github.com/anonymous-rep-exo/ATP_muscle_controller}{\texttt{ATP\_muscle\_controller}}.
\end{abstract}

\section{Introduction}

Exoskeletons have found widespread applications across industry~\cite{hypershell_official}, rehabilitation\cite{dragusanu2022design}, and the military~\cite{zoss2006biomechanical}. As the technology continues to evolve, bio-inspired approaches~\cite{gordon2022human} have emerged as particularly promising, offering comprehensive assistance by leveraging anatomical information. These methods, rooted in an understanding of human anatomy, have demonstrated significant potential to reduce the energy cost of movement.
In particular, lower-limb exoskeletons benefit greatly from anatomical assistance, which enables effective support across a wide range of tasks~\cite{molinaro2024task}. Anatomical assistance overcomes the limitations of exoskeleton-centered approaches, which often treat the human body as a black box or rely solely on external human--robot interaction forces as control interfaces.
This paradigm shift allows for assistance that is tailored to the principle of assist-as-needed.
By incorporating anatomical assistance, exoskeletons not only become more intelligent but also pave the way for generalized assistance in daily life.

\begin{figure}[!t]
    \centering
    \includegraphics[width=1.0\linewidth]{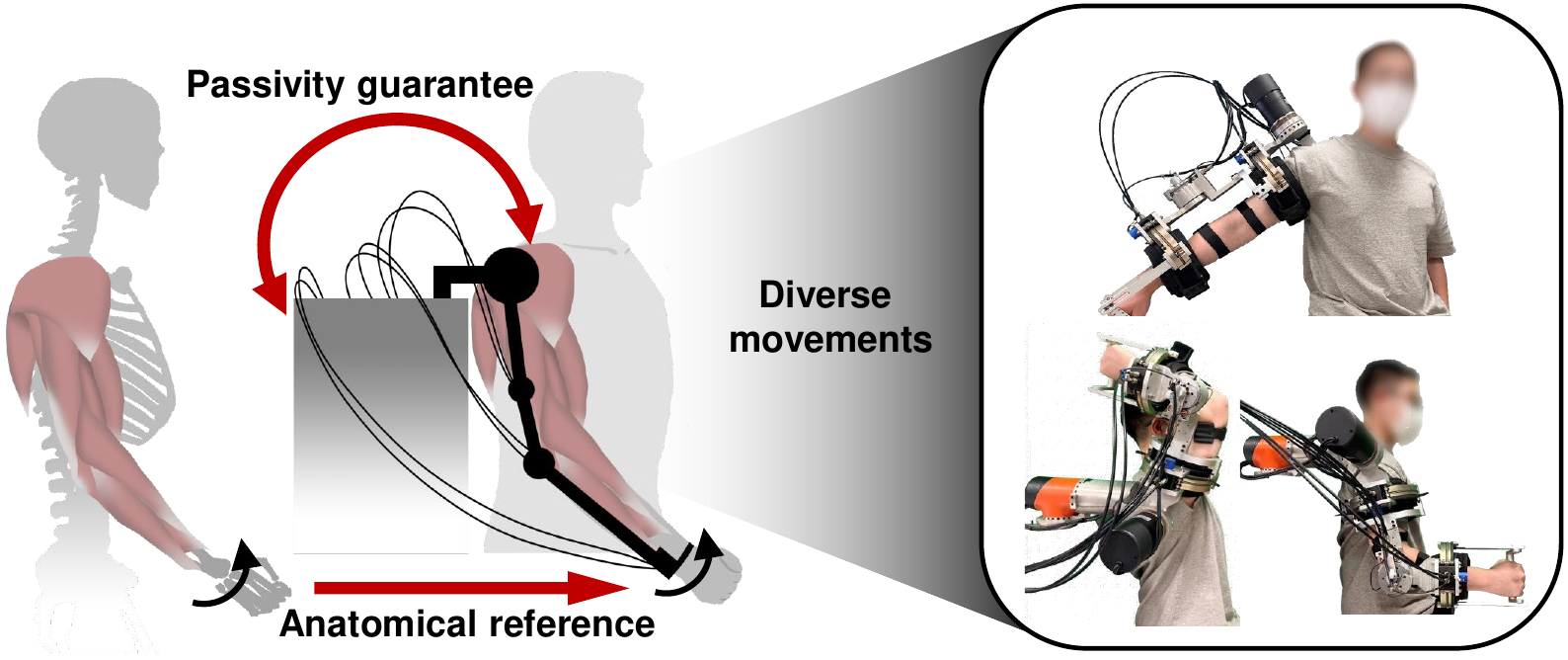}
    \caption{\textbf{Illustration of the anatomical assistance with exoskeleton.} The proposed method enables the upper-limb exoskeleton to provide anatomical assistance across diverse upper-limb movements.}
    \label{head}
    \vspace{-0.65cm}
\end{figure}

However, current anatomical assistance primarily focuses on lower-limb exoskeletons. This is largely because lower-limb exoskeletons are designed for weight support and involve periodic assistance~\cite{li2019human}, where the precision of support is less critical. 
In contrast, upper-limb exoskeletons face greater complexity in assisting limb movements, as they must account for both the complexity of upper-limb motions and the variability in user intentions~\cite{wang2025high,chen2026semantic}.
In assistance for diverse upper-limb movements, position-based regulation overly constrains limb freedom, while their nonperiodic and intention-dependent nature makes predefined assistance impractical, requiring anatomical assistance to be generated and delivered as torque profiles in real time from instantaneous interaction feedback.

Because human limb motion is generated through coordinated joint torques, musculoskeletal models provide a natural basis for deriving anatomical torque references that align exoskeleton assistance with human biomechanics without requiring task labels or predefined trajectories (Fig.~\ref{head}).
However, developing a unified muscle controller for complex musculoskeletal systems remains challenging. Moreover, the raw torque reference is unsuitable for direct deployment due to tendon-induced torque spikes, while the safety of human--robot interaction is not theoretically guaranteed. 
To address these challenges and provide versatile assistance based on anatomical knowledge of upper-limb exoskeletons, we propose the Anatomical Torque with Passivity-based Control (ATP) framework. 
To the best of our knowledge, this is the first framework for exoskeletons that generates anatomical assistance for multi-joint, complex, nonperiodic upper-limb movements without relying on predefined motion trajectories.
The contributions of this work can be summarized as follows:

\begin{itemize}
\item[-] A simulation environment for motion imitation in the musculoskeletal system is developed using the MuJoCo Playground~\cite{zakka2025mujoco} and MyoSim platforms~\cite{wang2022myosim}. This environment has been open-sourced\footnote{The mjlab version~\cite{zakka2026mjlab}, which supports manager-based training configuration, is also included in the repository.} to facilitate high-fidelity muscle--tendon simulation and large-scale training within the research community.

\item[-] A unified muscle controller based on reinforcement learning (RL) is developed to coordinate muscle activations within the musculoskeletal system and generate reference torques for a wide range of upper-limb movements. 
By incorporating anatomical knowledge, the controller generates torque references in real time without requiring complex biomechanical computations.

\item[-] An efficient online refinement method is introduced to improve the robustness and consistency of the anatomical reference torque. This method mitigates spiking behaviors caused by nonlinear tendon dynamics and enhances interaction safety. Additionally, the torque refinement process accounts for user comfort by adapting the assistance according to the anomaly score.

\item[-] A human--robot interaction torque controller based on the backstepping method is proposed for our cable-driven compliant upper-limb exoskeleton. Interaction safety is further enhanced by incorporating energy-tank technology, which preserves the passivity of the overall system. Rigorous theoretical proofs are provided to guarantee closed-loop torque-tracking performance as passivity is ensured. Simulation and real-world experiments are conducted to validate the proposed controller.

\end{itemize}

In this framework, the anatomical torque reference is learned from a simulated musculoskeletal system and refined online for practical delivery, while interaction safety is ensured through passivity.
We emphasize that torque-based assistance is particularly important for upper-limb movements, as it provides physical support without imposing predefined trajectories, thereby preserving the wearer’s motion freedom and accommodating complex, nonperiodic, and intention-dependent behaviors that are essential for safe and effective manipulation.


\section{Related Works}

This section overviews the research on assistance generation and interaction control for exoskeleton robots.\\

\noindent\textbf{Assistance Generation}:
Exoskeletons are designed to assist users in movement, operating on the principle of ``assist-as-needed''~\cite{teramae2017emg}, and recent advancements have demonstrated their potential by incorporating anatomical knowledge, such as metabolic rate~\cite{franks2022effects} and biological moments~\cite{bishe2022low}. Human-in-the-loop (HIL) optimization facilitates this by implicitly modeling the metabolic cost through sampling-based nonlinear optimization~\cite{HIL_first}, demonstrating effectiveness across various lower-limb exoskeletons, including prostheses~\cite{wen2017automatically}, soft exosuits~\cite{ding2018human}, and ankle exoskeletons~\cite{slade2022personalizing}. The optimized outcomes have also been validated across a range of walking tasks~\cite{chen2024learning}.
To address the time-consuming nature of online iterative stages, some researchers have turned to musculoskeletal modeling software, such as OpenSim~\cite{seth2018opensim}, which allows for closed-form calculations of metabolic rate. In~\cite{gordon2022human}, the musculoskeletal system is augmented with a hip exoskeleton to estimate bio-related costs in real time, thereby reducing the duration of HIL optimization. Additionally, in~\cite{molinaro2022subject}, joint moments are calculated offline by collecting joint motion and plantar pressure data. Furthermore, this moment estimator can be extended to various walking tasks through the use of temporal convolutional networks~\cite{molinaro2024task}.
However, existing methods often rely on metabolic measurement devices or complex biomechanical computations, primarily focusing on anatomical references for lower-limb motions and struggling to adapt to the upper-limb scenario, where movement patterns are more complex and lack clear periodic characteristics. Our framework aims to fill this gap by learning the anatomical torque in upper-limb movements directly, rather than measuring the ground-truth joint torque, and incorporating offline learning and online refinement, thereby enhancing practical applicability.

The growing demand for assisting humans in various daily tasks necessitates the development of exoskeletons with improved generalizability. RL has proven highly effective in augmenting and rehabilitating users across a broad range of applications~\cite{sharifi2025reinforcement}. In~\cite{rose2020end}, an end-to-end exoskeleton controller is trained offline by integrating the exoskeleton with a musculoskeletal model. By learning motion imitation behaviors and muscle coordination, the assistive controller for the exoskeleton can be concurrently trained within the same environment~\cite{luo2024experiment}, promoting generalization across diverse walking tasks.
Furthermore, explicit modeling of muscle activity~\cite{luo2023robust} and the application of policy distillation~\cite{luo2025robust} enhance the practicality of the exoskeleton, enabling its use in non-periodic movements such as sit-to-stand transitions~\cite{ratnakumar2024predicting}. 
Moreover, RL has been successfully applied to simpler upper-limb tasks, such as reaching~\cite{wannawas2024controlling}.
However, most existing approaches for learning anatomical reference torque rely on musculoskeletal models built in DART~\cite{lee2019scalable}, which lacks substantial support for large-scale parallel training and does not fully accommodate more detailed musculoskeletal models~\cite{zuo2024self, wang2022myosim}.
While progress has been made in GPU-accelerated training for musculoskeletal models~\cite{Li2026MuscleMimic}, the use of high-fidelity muscle--tendon simulations across different upper-limb tasks to deliver anatomical assistance remains unexplored.

\noindent\textbf{Interaction Control}: 
Interactive controllers must be carefully designed to meet both the assistance and safety requirements. The assistance provided by exoskeletons can be categorized into two distinct control schemes: trajectory-based~\cite{amiri2025fuzzy} and torque profile-based control~\cite{li2013semg}.
In the trajectory-based approach, impedance control aligns with the intrinsic nature of human interaction with different environments and tasks~\cite{cheah1998learning}. This has been extensively applied in robotic systems involving physical human--robot interaction. To adapt to varying interactive environments, variable impedance can be employed based on misalignment between the exoskeleton and the user~\cite{huo2021intention}. This approach can also be extended to an adjustment mechanism driven by a generative anomaly score, where human--robot conflicts are detected and mitigated by tuning the impedance~\cite{chen2024safe}. However, due to their reliance on predefined trajectories, these methods are limited to specific tasks and lack flexibility.
In contrast, assistance based on the torque profile is more tolerant of movement uncertainty, allowing greater flexibility for user movement~\cite{molinaro2024estimating,slade2024human}. However, this approach is less effective in complex upper-limb motions due to the loss of detail from the simplified trajectory. Finding a suitable method for complex assistive tasks that preserves detail while maintaining flexibility remains an open challenge.

Given the close and frequent nature of human--robot interactions in exoskeletons, safety is critical. Passivity provides a theoretical foundation for ensuring safe physical human--robot interaction~\cite{desoer2009feedback}, focusing on behaviors that prevent energy leakage and mitigate the risk of unexpected oscillations. In~\cite{zhang2015passivity}, passivity is ensured in a closed-loop upper-limb robotic system during mode transitions. In~\cite {spyrakos2019passivity}, passivity is achieved in a variable impedance control scheme in the presence of a compliant robot. Additionally, a tank-based approach~\cite{haddadin2024unified} has been developed by augmenting the impedance controller, enabling unified force and position control while considering passivity.

Although passivity preservation in robots has made significant progress, current approaches typically require predefined motion trajectories and lack torque-level implementations for cable-driven compliant actuators, limiting their applicability to flexible and safe exoskeleton assistance.

\section{Preliminaries}
\label{preli}
\begin{figure}[!t]
    \centering
    \includegraphics[width=0.66\linewidth]{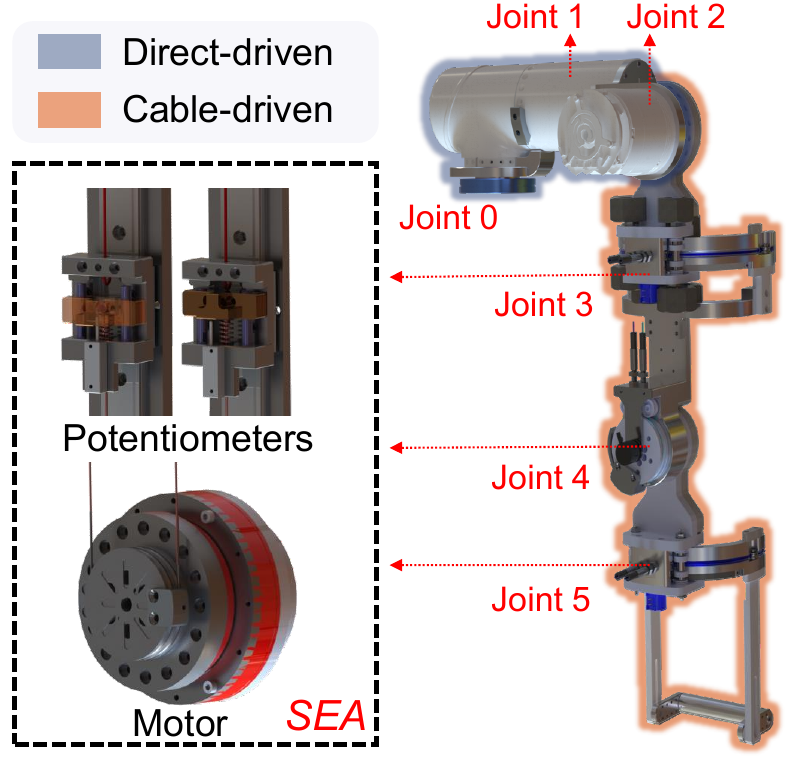}
    \caption{\textbf{Upper-limb exoskeleton structure} with direct-drive joint modules for Joints 1 and 2, and cable-driven SEA actuation for Joints 3 to 5.}
    \label{fig:mech}
\end{figure}

This study utilizes a cable-driven exoskeleton to provide assistance, with the mechanical structure depicted in Fig.~\ref{fig:mech}. The base (Joint 0) is a passive joint responsible for eccentric shoulder movement. Joints 1 and 2 are direct-drive modules responsible for shoulder abduction/adduction and shoulder flexion/extension, respectively. Joints 3 to 5 facilitate upper-arm internal and external rotation, elbow flexion/extension, and forearm internal and external rotation. These joints are cable-driven and incorporate series elastic actuators (SEAs), each unit consisting of a motor and two potentiometers, enabling the absorption of undesired impacts and ensuring backdrivability.

The dynamic model of the cable-driven upper-limb exoskeleton robot can be expressed as \cite{albu2007unified}
\begin{align}
\bm M(\bm q)\ddot{\bm q} + \bm C(\dot{\bm q}, \bm q)\dot{\bm q} + \bm g(\bm q) &= \bm S_1 \bm u + \bm S_2^{\mathsf{T}} \bm K (\bm \theta - \bm S_2 \bm q) + \notag \\
&\quad \bm \tau_{e} + \bm S_2^{\mathsf{T}} \bm \tau_{f}, \label{jointsys} \\
\bm B \ddot{\bm \theta} + \bm K (\bm \theta - \bm S_2 \bm q) &= \bm S_2 \bm u, \label{seasys}
\end{align}
using the notation detailed in Table~\ref{dynamic_table}. The dynamic parameters of the exoskeleton can be obtained from the datasheet or through the identification of the mechanical model. Additionally, the following conditions are imposed~\cite{cheah2015task,arimoto1996control}:
\begin{enumerate}
\item[1)] The matrix $\bm M(\bm q)$ is bounded, symmetric, and positive definite;
\item[2)] The term $\dot{\bm M}(\bm q) - 2 \bm C(\dot{\bm q},\bm q)$ is skew-symmetric;
\item[3)] The matrices $\bm B$ and $\bm K$ are constant, diagonal, and positive definite.
\end{enumerate}

\begin{table}[t]
\caption{Dynamic parameters}
\centering
\begin{tabular}{c|l}
\hline
$\bm M(\bm q)\in\mathbb{R}^{5\times 5}$ & Inertia matrix of robot \\ \hline
$\bm C(\dot{\bm q}, \bm q)\in\mathbb{R}^{5\times 5}$ & Matrix related to centripetal and Coriolis forces \\ \hline
$\bm g(\bm q)\in\mathbb{R}^5$ & Vector related to gravity \\ \hline
$\bm K\in\mathbb{R}^{3\times 3}$ & Stiffness matrix \\ \hline
$\bm q\in\mathbb{R}^5$ & Robot joint angles \\ \hline
$\bm B\in\mathbb{R}^{3\times 3}$ & Inertia matrix of motor \\ \hline
$\bm \theta\in\mathbb{R}^3$ & Motor rotation angles \\ \hline
$\bm\tau_e\in\mathbb{R}^5$ & Interaction torque with human participant \\ \hline
$\bm\tau_f\in\mathbb{R}^3$ & Disturbance torque \\ \hline
$\bm u\in\mathbb{R}^5$ & Control input exerted on robot joints \\ \hline
$\bm S_1 = \text{diag}(1, 1, 0, 0, 0)$ & Selection matrix for direct-driven joints \\ \hline
$\bm S_2 = [\bm 0, \bm I_3]$ & Selection matrix for cable-driven joints \\ \hline
\end{tabular}
\label{dynamic_table}
\vspace{-0.6cm}
\end{table}

\begin{figure*}[!t]
    \centering
  \includegraphics[width=0.9\linewidth]{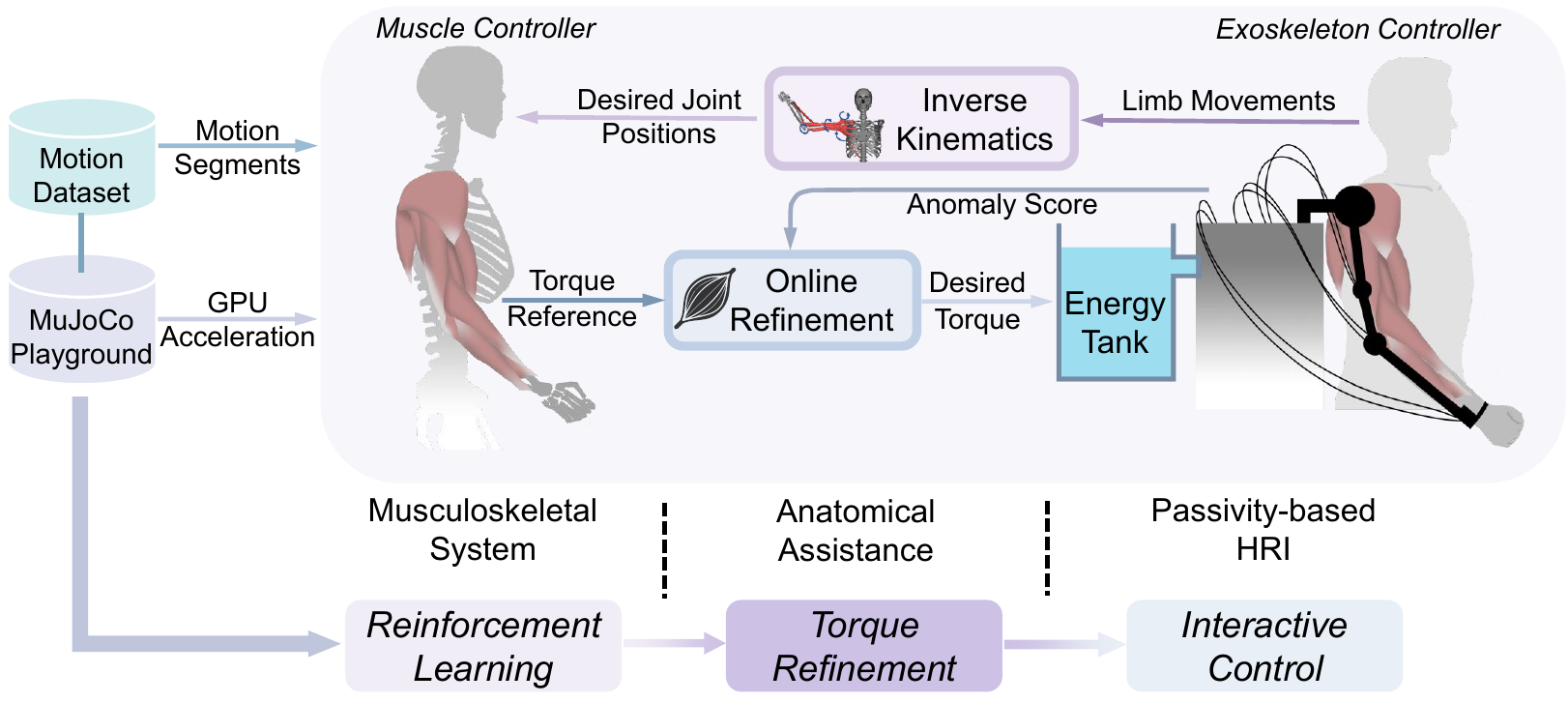}
    \caption{\textbf{The framework of ATP}: A muscle controller with anatomical knowledge is rapidly trained on a GPU-accelerated musculoskeletal system by using RL. The real-time generated joint reference torque is refined online by incorporating safety constraints and an anomaly score with comfort information, serving as the desired human--robot interaction torque for the exoskeleton controller. Passivity of human--robot interaction is ensured through energy-tank integration, enabling safe anatomical assistance.}
    \label{Fig:overallFigure}
\end{figure*}
\begin{table*}[t!]
\caption{Comparison of Simulation Environments}
\begin{tabular}{|l|l|l|l|l|l|}
\hline
\textbf{Environment} & \textbf{Physics Engine} & \textbf{Implementation} & \textbf{Supported Model Formats} & \textbf{GPU Acceleration} & \textbf{Tendon Modeling} \\ \hline
\cellcolor[HTML]{F1B4AE} IsaacLab\textsuperscript{\textcolor[HTML]{F1B4AE}{1}}    & PhysX         & C++  & USD, URDF    & $\surd$   & Custom \\ \hline
\cellcolor[HTML]{CCEBA5} Newton\textsuperscript{\textcolor[HTML]{CCEBA5}{2}}       & Newton & Python  & USD, MJCF, URDF  & $\surd$   & Custom \\ \hline
\cellcolor[HTML]{B3CDEE} MuJoCo\textsuperscript{\textcolor[HTML]{B3CDEE}{3}}       & MuJoCo        & Python  & MJCF   & -    & Tendon Available \\ \hline
\cellcolor[HTML]{B3CDEE} MuJoCo Playground\textsuperscript{\textcolor[HTML]{B3CDEE}{3}} & JAX MuJoCo    & Python  & MJCF   & $\surd$   & Tendon Available \\ \hline
DART             & DART (ODE/Bullet) & C++ & Custom Model Setup & -   & Custom \\ \hline
\cellcolor[HTML]{F1B4AE} Unity\textsuperscript{\textcolor[HTML]{F1B4AE}{1}}      & PhysX         & C++  & USD, FBX  & $\surd$   & Custom \\ \hline
\end{tabular}
\label{tab:env}
\vspace{0.1cm}

\begin{tablenotes}
\footnotesize
\item \textcolor[HTML]{F1B4AE}{Red} represents environments with the most realistic visual quality; \textcolor[HTML]{B3CDEE}{Blue} represents environments with the highest simulation accuracy for contact and collisions; \textcolor[HTML]{CCEBA5}{Green} represents environments that are differentiable.
\end{tablenotes}
\vspace{-0.5cm}
\end{table*}

The framework of this study is depicted in Fig.~\ref{Fig:overallFigure}. The muscle controller is trained within the MuJoCo Playground using a large-scale motion dataset, enabling the real-time simulation of muscle activity based on limb movement. This process generates the joint torque, which is extracted as the torque reference, $\bm\tau_r$. The torque refinement module adjusts the torque reference by incorporating safety constraints, as well as considering the comfort metric generated by the anomaly detector~\cite{chen2025upper}. The refined torque, which integrates anatomical knowledge and safety factors, is referred to as the desired torque, $\bm\tau_d$, and serves as the command input to the exoskeleton controller. This controller facilitates accurate regulation of the human--robot interaction torque while ensuring passivity through the integration of energy-tank technology. 
In summary, anatomical knowledge is efficiently embedded within the muscle controller and transferred to the exoskeleton controller, with safety considerations, to provide anatomical assistance for various upper-limb movements.\\

\section{Anatomical Assistance}
\label{Biomechanical_sec}
Closed-form joint-torque computation cannot meet the real-time requirements of the proposed framework, while learning-based approaches remain difficult to train because of the highly nonlinear musculoskeletal system and the high-dimensional control space. Moreover, directly applying the learned anatomical reference is impractical, as tendon dynamics and neural-network variability can introduce sharp torque spikes.
Therefore, this section presents a method for learning anatomical reference torques through reward shaping in parallel simulation environments. The learned references are then refined online using interaction feedback to generate safe and anatomical torque commands in real time.\\



\begin{figure}[!t]
    \centering
    \includegraphics[width=0.8\linewidth]{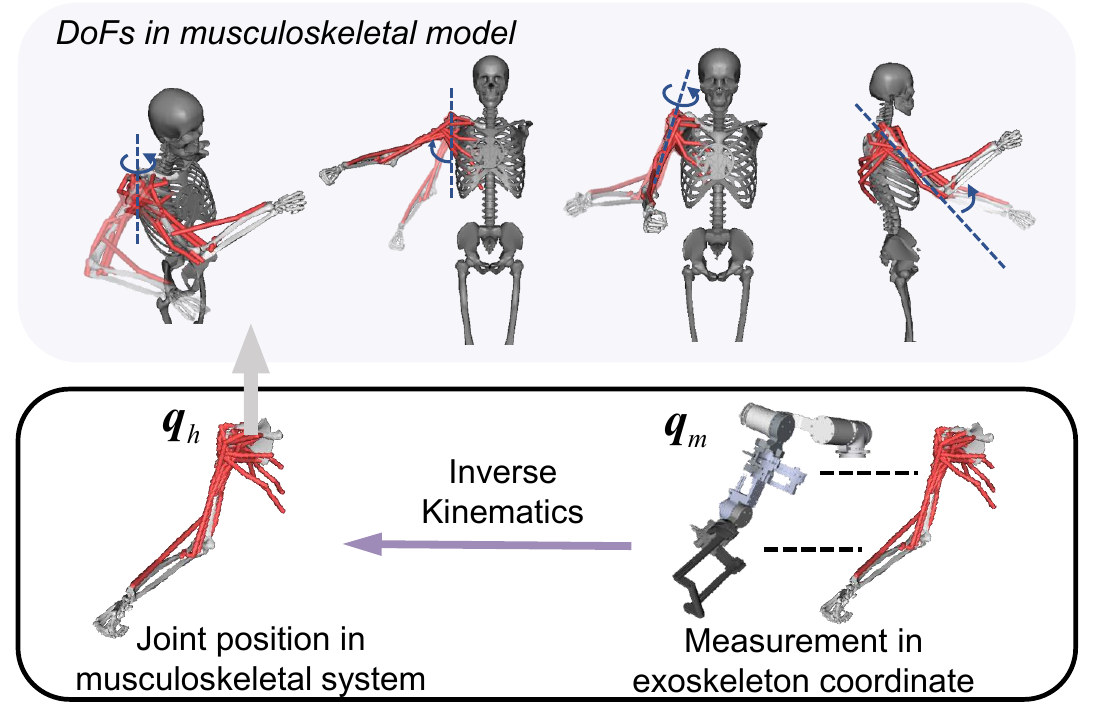}
    \caption{\textbf{Kinematic mapping between the exoskeleton and musculoskeletal model} for four-axis alignment, enabling delivered assistance to be transformed into the exoskeleton coordinate.}
    \label{fig:myomodel}
\end{figure}

\noindent\textbf{Anatomical Torque Learning}: 
To calculate the joint torque in real time based on limb movement, we developed an open-source musculoskeletal simulation environment within an existing simulation platform. It supports efficient training of a unified muscle controller across diverse upper-limb movements.
The characteristics of different simulation environments are summarized in Table~\ref{tab:env}. 
Considering the requirements for tendon simulation, training efficiency, and native support for MJCF musculoskeletal models, we adopted MuJoCo Playground and deployed the MyoArm upper-limb musculoskeletal model (Fig.~\ref{fig:myomodel}), which provides fast and physiologically accurate muscle simulation for controller training.
The tendons responsible for controlling wrist and hand movement are disabled, as the developed exoskeleton provides limited active assistance due to the loose connection to the wrist (Joint 5). 
This model consists of 24 muscle tendons with 15 degrees of freedom, including 4 independent degrees of freedom $\bm q_h\in\mathbb{R}^4 $ that drive the coordinated motion of other joints.

Since the corresponding joint axes differ between the musculoskeletal model and the exoskeleton, an inverse kinematic mapping from the measured exoskeleton coordinates $\bm q_m$ to the musculoskeletal coordinates $\bm q_h$ is established to ensure a consistent assistance effect at the wearer’s joints. 
Specifically, the joint positions $\bm q_h$ are computed as follows:
\begin{align}
    \bm q_h=&f^{-1}\left(\bm q_m\right),\\
    \dot{\bm q}_m=&\bm J\left(\bm q_h\right)\dot{\bm q}_h,
    \label{equ:jacobmh}
\end{align}
where $f(\cdot)$ represents the forward kinematics, and $\bm J\left(\cdot\right)$ is the Jacobian matrix that relates the musculoskeletal coordinates to the exoskeleton coordinates.

In this study, we model the muscle behavior for reference trajectory tracking as a Markov decision process, defined by the tuple $\mathcal{M} = \langle \mathcal{S}, \mathcal{A}, \mathcal{P}, \mathcal{R}, \gamma \rangle$, where $\mathcal{S}$ represents the state space, $\mathcal{A}$ the action space, $\mathcal{P}$ the state transition function, $\mathcal{R}$ the reward function, and $\gamma$ the discount factor. To facilitate RL for anatomical assistance, the action $\mathbf a_t \in \mathcal{A}$ is defined as a 24-dimensional vector, where each element is in the range of $ [0, 1]$, representing the activation intensity of the tendon at each timestep $t$.
The observed state $\mathbf s_t \in \mathcal{S}$ is described by
\begin{align}
    \mathbf s_t=&\left(\mathbf p,\dot{\mathbf p},{\mathbf p}_r,\dot{\mathbf p}_r,\mathbf e_p,{\mathbf e}_{\dot{p}},\bm l_t, \dot{\bm l}_t,\mathbf f_t\right),
    \label{eqn:RLstate}
\end{align}
where $\mathbf p \in \mathbb{R}^{15}$ denotes the pose of the musculoskeletal model, $\mathbf p_r \in \mathbb{R}^{15}$ represents the reference pose, $\mathbf e_p \triangleq \mathbf p - \mathbf p_r$ and $\mathbf e_{\dot{p}} \triangleq \dot{\mathbf p} - \dot{\mathbf p}_r$ are the tracking position and velocity errors, and $\bm l_t, \dot{\bm l}_t, \mathbf f_t$ represent the tendon length, tendon velocity, and tendon force, respectively.

We apply the proximal policy optimization (PPO) algorithm to train the muscle controller, aiming to maximize the cumulative discounted reward $\mathbb{E}\left[\sum_t\gamma^{t-1}r_t\right]$. 
To incorporate anatomical knowledge into the RL process, the following reward function is defined:
\begin{align}
r = &r_{\text{track}} + r_{\text{energy}}, \\
     r_{\text{track}} =&
    \exp\left(-\frac{\operatorname{mean}\left(\mathbf{e}_p^{\circ 2}\right)}{\sigma_p}\right)                 
    + \exp\left(-\frac{\operatorname{mean}\left(\mathbf{e}_{\dot{p}}^{\circ 2}\right)}{\sigma_{\dot{p}}}   
  \right), \\
  r_{\text{energy}}=&-\lambda_a \Vert\mathbf{a}_t\Vert, 
\end{align}
where $\lambda_a$ is a scale constant, and $(\cdot)^{\circ 2}$ refers to the element-wise square. Reward shaping~\cite{xie2025kungfubot} is introduced to accelerate and enhance the robustness of training on long-duration motions. Weights $\sigma_p$ and $\sigma_{\dot{p}}$ are updated at each state transition as follows:
\begin{align}
    \bar{e}_p &\leftarrow (1 - \alpha_{p})\bar{e}_p+ \alpha_p \cdot \operatorname{mean}\left(\mathbf{e}_p^{\circ 2}\right), \\
    \bar{e}_{\dot{p}} &\leftarrow (1 - \alpha_{\dot{p}})\bar{e}_{\dot{p}}+ \alpha_{\dot{p}} \cdot \operatorname{mean}\left(\mathbf{e}_{\dot{p}}^{\circ 2}\right), \\
    \sigma_p &\leftarrow\min(\sigma_p,\bar{e}_p),\\
    \sigma_{\dot{p}} &\leftarrow\min(\sigma_{\dot{p}},\bar{e}_{\dot{p}}),
\end{align}
where $\alpha_p$ and $\alpha_{\dot{p}}$ are constants that govern the memory length of the adaptive weights. Through offline parallel training within the musculoskeletal system, the muscle controller $\pi\left(\mathbf a_t|\mathbf s_t\right)$ captures the anatomical behavior. The joint torque reference in the exoskeleton coordinate is obtained as follows:
\begin{align}
    \bm\tau_r=&\bm J^{-T}\left(\bm q_h\right)\bm\tau_h,\\
    \label{equ:tau_ref}
    \bm\tau_h=&f_{\pi}\left(\bm q_m\right),
\end{align}
where $f_{\pi}(\cdot)$ represents the musculoskeletal model driven by the trained muscle controller, which generates the corresponding joint torques $\bm\tau_h$ for the independent joints $\bm q_h$ required to achieve trajectory tracking, based on the reference pose $\mathbf{p}_r$ calculated from $\bm q_m$. By deploying the effectively trained muscle controller on the musculoskeletal model, the anatomical torque reference can be obtained in real time.

\noindent\textbf{Online Torque Refinement}:
Due to the closeness and frequency of human--robot interactions in the exoskeleton, safety requirements are a crucial consideration in anatomical assistance. To quantify the human--robot interactions, a diffusion model-based approach is applied~\cite{chen2025upper} to generate an anomaly score:
\begin{align}
    s=f_a\left(\bm q, \dot{\bm q}, \bm \theta, \dot{\bm \theta},\bm \tau_e\right),
    \label{equ:anomaly_score}
\end{align}
where $f_a(\cdot)$ represents the anomaly detector, which is trained on interactive data collected from healthy subjects. The subjects move the upper-limb exoskeleton robot at their preferred speed, whether rapidly, slowly, or in a stationary position, while the exoskeleton operates in transparent mode. By extracting the multi-modal data distribution patterns from normal interactions, the anomaly detector identifies various anomalies and generates a score in which lower values signal higher comfort.

Since the reference torque is generated directly by simulation, the muscle controller may exhibit unexpected torque spikes during trajectory tracking due to the inherent stochasticity of neural networks and tendon dynamics. To incorporate safety considerations and ensure stable assistance, the following online torque refinement is proposed.
\begin{align}
\label{opt_all}
\min_{\bm \tau_d, \bm u_d, s} &\sum_{i=t}^{t+N_p}\left[\|\bm \tau_{d}^{(i)} - \bm \tau_{r}^{(i)}\|_{\bm Q}^2 + \|\bm u_{d}^{(i)}\|_{\bm R}^2 + (s^{(i)})^2\right],\\
 s.t.\quad &{\bm \tau}_d^{(t+1)} = \bm \tau_d^{(t)} + \bm u_d^{(t)}\Delta t\notag\\
 &s^{(t+1)}=s^{(t)} + (\frac{\partial f_a}{\partial \bm \tau_e})^{\mathsf{T}}\bm u_d^{(t)}\Delta t\notag\\
&\bm \tau_d \in \mathcal{T}, \bm u_d \in \mathcal{U} \notag
\end{align}
where $N_p$ is the prediction horizon, $\Delta t$ is the time interval, $\bm u_d$ is the rate of change of the desired torque, $\bm Q$ and $\bm R$ are symmetric positive-definite weighting matrices, the superscript denotes the time step, and $\mathcal{T}$ and $\mathcal{U}$ represent the sets of feasible torque commands and their rate of change, respectively, ensuring the assistance remains within a safe region while improving comfort.
Note that the desired interaction torque on the robot side is set to $-\bm{\tau}_d$ to provide the corresponding assistive torque $\bm{\tau}_d$ to the wearer.\\

\section{Interactive control}
\label{inter_sec}
Online torque refinement generates safe interaction torque commands informed by anatomical knowledge. 
However, directly tracking this torque command does not guarantee safe physical human–robot interaction, especially for upper-limb exoskeletons involving frequent contact and bidirectional energy exchange.
This section presents a passivity-based controller for the cable-driven SEA exoskeleton that enables accurate anatomical torque tracking while providing a theoretical guarantee of interaction safety.\\

\noindent\textbf{Backstepping Torque Control}: Given that the desired assistance profile $\bm\tau_d$ is obtained via biomechanical analysis and dynamically constrained refinement, the exoskeleton is required to accurately deliver the assistance and ensure interaction safety in real time.
To this end, the backstepping method\cite{li2016adaptive} is employed to construct the torque-tracking controller. 
For the cable-driven joint, the desired joint position for the motor is designed as
\begin{align}
    \bm \theta_d=&\bm S_2\bm q+\bm K^{-1}\bigl\{\bigl[-k_g\bm{sgn}(\dot{\tilde{\bm q}})-\hat{\bm\tau}_f\big]+\bm S_2\big[\bm M(\bm q)\ddot{\hat{\bm q}}\notag \\
    &+\bm C(\dot{\bm q}, \bm q)\dot{\hat{\bm q}}+\bm g(\bm q)-\bm K_v\dot{\tilde{\bm q}}-\bm F_f\bigr]\big\},\label{equ:desire_th}\\
    \bm F_f=&\bm K_p\tilde{\bm\tau}+\bm K_d\dot{\tilde{\bm\tau}}+\bm\tau_e,
\end{align}
where $\bm K_v,\bm K_p,\bm K_d\in \Re^{5\times5}$ are constant gain matrices, $\hat{\bm\tau}_f$ is the friction estimated offline via polynomial fitting\cite{chen2025upper}, and $k_g$ is a constant, $\bm F_f$ is a torque tracking-related term, the torque-tracking error is ${\tilde{\bm\tau}}\triangleq \bm\tau_d-\bm\tau_e$, and ${\tilde{\bm q}}\triangleq\bm q -\hat{\bm q}$ is the position-tracking error, where the command $\hat{\bm q}$ is derived from the following joint observer:
\begin{eqnarray}
    \left \{
    \begin{array}{*{20}{l}}
    \dot{\hat{\bm q}} = \bm\eta_q+\beta_q\tilde{\bm q}\\
    \dot{\bm\eta}_q=\bm M(\bm q)^{-1}\bigl[\bm S_1\bm u+\bm S_2^{\mathsf{T}}\bm K(\bm \theta-\bm S_2\bm q)\\
    \qquad+\bm\tau_e+\bm S_2^{\mathsf{T}}\hat{\bm\tau}_f-\bm C(\dot{\bm q}, \bm q){\bm\eta}_q-\bm g(\bm q)+\bm K_q\tilde{\bm q}\bigr]
    \end{array} \right..
\label{equ:joint_ob}
\end{eqnarray}
Here, $\beta_q$ and $\bm K_q$ are a constant and a constant gain matrix, respectively, and $\bm\eta_q$ is an auxiliary variable. Following the same principle, the controller input for the direct-driven joint is set as
\begin{align}
    \bm S_1\bm u=\bm S_1[\bm M(\bm q)\ddot{\hat{\bm q}}+\bm C(\dot{\bm q}, \bm q)\dot{\hat{\bm q}}+\bm g(\bm q)-\bm K_v\dot{\tilde{\bm q}}-\bm F_f]\label{equ:u_direct}.
\end{align}

Substituting (\ref{equ:desire_th}) and (\ref{equ:u_direct}) into (\ref{jointsys}), one obtains the closed-loop equation on the joint side:
\begin{align}
    \bm M(\bm q)\ddot{\tilde{\bm q}}&+\bm C(\dot{\bm q}, \bm q)\dot{\tilde{\bm q}}+\bm K_v\dot{\tilde{\bm q}}-\bm S_2^{\mathsf{T}}\bm K\Delta\bm\theta\notag\\
    &+\bm S_2^{\mathsf{T}}\bigl(k_g\bm{sgn}(\dot{\tilde{\bm q}})-\tilde{\bm\tau}_f\bigr)+\bm F_f-\bm\tau_e=0,
    \label{equ:dynamic2}
\end{align}
where $\tilde{\bm\tau}_f={\bm\tau}_f-\hat{\bm\tau}_f$ and $\Delta\bm\theta=\bm\theta-\bm\theta_d$ are the friction estimation error and the position-tracking error on the motor side, respectively.
The following Proposition can be derived to guarantee the convergence of $\tilde{\bm \tau}$.

\noindent\textbf{Proposition 1}:
{\em In the presence of $\Vert \tilde{\bm\tau}_f \Vert \hspace{-0.05cm}\leq\hspace{-0.05cm} \zeta$ where $\zeta$ denotes the bound, and $\dot{\tilde{\bm q}}\rightarrow \bm0$ and $\Delta\bm\theta\rightarrow \bm0$ are satisfied, the torque-tracking error ${\tilde{\bm\tau}}$ is uniformly ultimately bounded.}
\begin{IEEEproof}
Subtracting (\ref{equ:joint_ob}) from (\ref{jointsys}) yields
\begin{align}
    \bm M(\bm q)\ddot{\tilde{\bm q}}&+\big(\bm C(\dot{\bm q}, \bm q)+\beta_q\bm M(\bm q)\big)\dot{\tilde{\bm q}}\notag \\ 
    &+\big(\beta_q\bm C(\dot{\bm q}, \bm q)+\bm K_q\big)\tilde{\bm q}=\bm S_2^{\mathsf T}\tilde{\bm\tau}_f, \label{equ:initial_res}
\end{align}
By choosing a sufficiently large $\beta_q$ and a correspondingly larger $\bm K_q$, there exists a finite $\mu$ such that $\limsup_{t\rightarrow\infty}\Vert\bm M(\bm q)\ddot{\tilde{\bm q}}\Vert\leq\mu$ holds whenever $\Vert \tilde{\bm\tau}_f \Vert \hspace{-0.05cm}\leq\hspace{-0.05cm} \zeta$.

Consider the following candidate Lyapunov function:
\begin{eqnarray}
&V_e = \frac{1}{2}\Tilde{\bm\tau}^{\mathsf{T}}\bm K_d\Tilde{\bm\tau}
\label{Lya_tau}.
\end{eqnarray}
Differentiating $V_e$ with respect to time and substituting (\ref{equ:dynamic2}) yields
\begin{align}
\dot V_e = & -\Tilde{{\bm\tau}}^{\mathsf{T}}\bm K_p\Tilde{{\bm\tau}} \notag \\
& - \tilde{\bm\tau}^{\mathsf{T}}\big(\bm M(\bm q)\ddot{\tilde{\bm q}}+\bm C(\dot{\bm q}, \bm q)\dot{\tilde{\bm q}}+\bm K_v\dot{\tilde{\bm q}}-\bm S_2^{\mathsf{T}}\bm K\Delta\bm\theta\big)\notag \\
&-\Tilde{{\bm\tau}}^{\mathsf{T}}\bm S_2^{\mathsf{T}}\bigl(k_g\bm{sgn}(\dot{\tilde{\bm q}})-\tilde{\bm\tau}_f\bigr)
\label{dot_Lya_tau}.
\end{align}

In the presence of $\dot{\tilde{\bm q}}\rightarrow \bm0$ and $\Delta\bm\theta\rightarrow \bm0$, the following can be deduced according to the Schwarz inequality and the boundedness of $\Vert \tilde{\bm\tau}_f \Vert$ and $\Vert\bm M(\bm q)\ddot{\tilde{\bm q}}\Vert$:
\begin{align}
\dot V_e \leq& -\lambda_{\min}[\bm K_p]\Vert\Tilde{{\bm\tau}} \Vert^2 + (\sqrt{3}k_g+\zeta+\mu)\Vert\Tilde{{\bm\tau}} \Vert \notag \\
=& -(1-\varrho)\lambda_{\min}[\bm K_p]\Vert\Tilde{{\bm\tau}} \Vert^2 - \varrho\lambda_{\min}[\bm K_p]\Vert\Tilde{{\bm\tau}} \Vert^2 \notag \\
& + (\sqrt{3}k_g+\zeta+\mu)\Vert\Tilde{{\bm\tau}} \Vert
\label{dot_Lya_tau1},
\end{align}
where $\varrho \in (0,1)$, and $\lambda_{max}[\cdot],\lambda_{min}[\cdot]$ represent the maximum and the minimum eigenvalues. 

Since $V_e \leq \frac{1}{2}\lambda_{\max}[\bm K_d]\Vert\Tilde{{\bm\tau}} \Vert^2$, the inequality of
\begin{align}
\dot V_e \leq& -\frac{2(1-\varrho)\lambda_{\min}[\bm K_p]}{\lambda_{\max}[\bm K_d]} V_e
\label{dot_Lya_tau3}
\end{align}
is obtained if
\begin{align}
\Vert\Tilde{{\bm\tau}} \Vert \geq &\frac{\sqrt{3}k_g+\zeta+\mu}{\varrho\lambda_{\min}[\bm K_p]} 
\label{dot_Lya_tau2}.
\end{align}

Here, (\ref{dot_Lya_tau2}) denotes the condition of negativity of the last two terms of (\ref{dot_Lya_tau1}), and further guarantees the inequality (\ref{dot_Lya_tau3}). 

Therefore, $\Tilde{{\bm\tau}}$ converges asymptotically to the ball with radius $(\sqrt{3}k_g+\zeta+\mu)/{\varrho\lambda_{\min}[\bm K_p]}$\cite{arimoto1996control}.
\end{IEEEproof}

Note that the role of the joint observer is two-fold: first, it eliminates the need for second-order information of the joint angle; and second, it uses the observed angle as the desired position at the joint end, thereby simplifying the problem into a torque-tracking problem.

For the motor side of the cable-driven SEA joints, the following controller is proposed to track the desired position and further realize the torque tracking
\begin{align}
    \bm S_2\bm u&=\bm K(\bm\theta-\bm S_2\bm q)+\bm B\ddot{\bm\theta}_r-\bm K_s\bm s-\bm K_{\theta}\Delta\hat{\bm\theta}\label{equ:u_sea},\\
    \bm s&=\dot{\bm\theta}-\dot{\bm\theta}_r,\\
    \dot{\bm\theta}_r&=\dot{\hat{\bm\theta}}_d-\alpha_{\theta}\Delta\hat{\bm\theta},
\end{align}
where $\bm K_s,\bm K_{\theta}$ are constant matrices, $\bm s$ is a sliding vector defined with a reference vector $\dot{\bm\theta}_r$, and $\alpha_{\theta}$ is a constant. Specifically, the position-tracking error is reformulated as $\Delta\hat{\bm\theta}=\bm\theta-\hat{\bm\theta}_d$, where $\hat{\bm\theta}_d$ denotes an estimated desired input, which is used to eliminate the requirement for higher-order information on the joint side during torque tracking~\cite{liu2008adaptive}. The observer is constructed as
\begin{eqnarray}
    \left \{
    \begin{array}{*{20}{l}}
    \dot{\hat{\bm \theta}}_d = \bm\eta_e+\beta_e\bm e\\
    \dot{\bm\eta}_e=\bm B^{-1}[\bm S_2\bm u-\bm K(\bm \theta-\bm S_2\bm q)+\bm K_e\bm e]
    \end{array} \right.,
\label{equ:joint_des_ob}
\end{eqnarray}
where $\beta_e,\bm K_e$ are a constant parameter and a constant matrix, respectively, $\bm e \triangleq\bm\theta_d-{\hat{\bm \theta}}_d$ is the observation error, and $\bm\eta_e$ is an auxiliary variable. 
Substituting (\ref{equ:u_sea}) into (\ref{seasys}) yields the closed-loop equation on the actuator side:
\begin{align}
    \bm B\dot{\bm s}+\bm K_s\bm s+\bm K_{\theta}\Delta\hat{\bm\theta}=0.
    \label{equ:closed-loop2}
\end{align}

It can be proved that the closed-loop system on the actuator side asymptotically converges to the origin. Once the convergence to the joint command $\hat{\bm q}$ is guaranteed, the torque-tracking controller described by (\ref{equ:desire_th}), (\ref{equ:u_direct}), and (\ref{equ:u_sea}) can achieve torque tracking with a known accuracy according to Theorem 1.\\

\noindent\textbf{Passivity-Based Correction}: Since the assistance is delivered by an exoskeleton closely attached to the wearer, safety must be assured prior to interactive torque tracking. That is, passivity must be confirmed before any delivery of assistance. The following storage function describes the energy of the robot system:
\begin{align}
    H=&\frac{1}{2}\dot{\tilde{\bm q}}^{\mathsf{T}}\bm M(\bm q)\dot{\tilde{\bm q}}+\frac{1}{2}\bm z^{\mathsf{T}}\bm B\bm z+\frac{1}{2}\bm s^{\mathsf{T}}\bm B\bm s\\\notag
    &+\frac{1}{2}\Delta\hat{\bm\theta}^{\mathsf{T}}(2\alpha_{\theta}\bm K_s+\bm K_{\theta})\Delta\hat{\bm\theta}
\end{align}
where $\bm z\triangleq \dot{\bm\theta}-\bm\eta_e$ is the estimated speed error.
Substituting (\ref{equ:dynamic2}) and (\ref{equ:closed-loop2}) into the time derivative of the storage function yields
\begin{align}
    \dot H=&\dot{\tilde{\bm q}}^{\mathsf{T}}[-\bm K_v\dot{\tilde{\bm q}}+\bm S_2^{\mathsf{T}}\bm K\Delta\bm\theta -\bm S_2^{\mathsf{T}}\bigl(k_g\bm{sgn}(\dot{\tilde{\bm q}})\notag \\
    &-\tilde{\bm\tau}_f\bigr)-\bm F_f+\bm\tau_e]-\bm z^{\mathsf{T}}\bm K_e \bm e-\bm s^{\mathsf{T}}\bm K_s\bm s\notag \\
    &-\bm s^{\mathsf{T}}\bm K_{\theta}\Delta\hat{\bm\theta}+\Delta\hat{\bm\theta}^{\mathsf{T}}(2\alpha_{\theta}\bm K_s+\bm K_{\theta})\Delta\dot{\hat{\bm\theta}}.
\end{align}
Using $\bm z=\Delta\dot{\hat{\bm\theta}}+\beta_e\bm e$, $\bm s=\Delta\dot{\hat{\bm\theta}}+\alpha_{\theta}\Delta\hat{\bm\theta}$ and $\Delta{\bm\theta}=\Delta\hat{\bm\theta}-\bm e$, this can be rewritten as
\begin{align}
    \dot H=&\dot{\tilde{\bm q}}^{\mathsf{T}}[-\bm F_f+\bm\tau_e]-\dot{\tilde{\bm q}}^{\mathsf{T}}\bm S_2^{\mathsf{T}}\bigl(k_g\bm{sgn}(\dot{\tilde{\bm q}})-\tilde{\bm\tau}_f\bigr) - \bm y^{\mathsf{T}} \bm P\bm y,
    \label{equ:Hdot}
\end{align}
where $\bm y=[\dot{\tilde{\bm q}}^{\mathsf{T}},\Delta{\hat{\bm\theta}}^{\mathsf{T}} ,\Delta\dot{\hat{\bm\theta}}^{\mathsf{T}},\bm e^{\mathsf{T}}]^{\mathsf{T}}$ and $\bm P$ is given as

\begin{equation}
\bm{P} =
\begin{bmatrix}
\bm K_v & -\bm S_2^{\mathsf{T}}\frac{\bm K}{2} & \bm 0 & \bm S_2^{\mathsf{T}}\frac{\bm K}{2}\\
-\frac{\bm K}{2}\bm S_2 & \alpha_{\theta}^2\bm K_s+\alpha_{\theta}\bm K_{\theta} & \bm0 & \bm0\\
\bm 0 & \bm 0 & \bm K_s & \frac{1}{2}\bm K_e \\
\frac{\bm K}{2}\bm S_2 & \bm 0 & \frac{1}{2}\bm K_e & \beta_e \bm K_e
\end{bmatrix}
\end{equation}
Without loss of generality, $\bm K_v$ is taken to be block-diagonal with respect to the direct-driven and cable-driven joint partition, and we let $\bar{\bm K}_v\triangleq\bm S_2\bm K_v\bm S_2^{\mathsf{T}}\in\mathbb{R}^{3\times 3}$ denote its restriction to the cable-driven SEA subspace. The parameters $\bm K_v,\bm K_s,\bm K_\theta$, and $\bm K_e$ are chosen such that
\begin{align}
    &\alpha_{\theta}^2\lambda_{min}[\bar{\bm K}_v\bm K_s]+\alpha_{\theta}\lambda_{min}[\bar{\bm K}_v\bm K_{\theta}]>\frac{1}{4}\lambda_{max}[\bm K^2]\label{equ:con1}\\
    &\beta_e\lambda_{min}[\bm K_s]>\frac{1}{4}\lambda_{max}[\bm K_e]\label{equ:con2}\\
    &\lambda_{min}[\bm K_e]\{\beta_e\lambda_{min}[\bm K_s]-\frac{1}{4}\lambda_{max}[\bm K_e]\}\notag\\
    &\times\{\alpha_{\theta}^2\lambda_{min}[\bar{\bm K}_v\bm K_s]+\alpha_{\theta}\lambda_{min}[\bar{\bm K}_v\bm K_{\theta}]-\frac{1}{4}\lambda_{max}[\bm K^2]\}\notag\\
    &>\frac{1}{4}\lambda_{max}[\bm K_s(\alpha_{\theta}^2\bm K_s+\alpha_{\theta}\bm K_{\theta})]\lambda_{max}[\bm K^2],
    \label{equ:con3}
\end{align}
where the matrix $\bm P$ is positive definite. 
Additionally, by setting $k_g>\zeta$, where $\zeta$ is the upper bound of the friction estimation error, the last two terms in (\ref{equ:Hdot}) are both made to be negative.

\begin{table*}[!ht]
\caption{Passivity guarantee of the time derivative of the storage function}
\centering
\begin{tblr}{
  cell{2}{4} = {r=4}{},
  hline{1,6} = {-}{0.08em},
  hline{2} = {-}{},
}
$\gamma_f$ & $\gamma_o$ & Tank-related terms $R$ & Passivity \\
0 & 0 &      $-\dot{\bm q}^{\mathsf{T}}\underbrace{\overline{\bm F}_f}_{=\alpha_f\bm F_f}+\alpha_f\dot{\bm q}^{\mathsf{T}}{\bm F}_f+\underbrace{\dot{\overline{\hat{\bm q}}}^{\mathsf{T}}}_{=\alpha_o\dot{{\hat{\bm q}}}^{\mathsf{T}}}(\overline{\bm F}_f-\bm\tau_e)-\alpha_o\dot{{\hat{\bm q}}}^{\mathsf{T}}(\overline{\bm F}_f-\bm\tau_e)=0$            &     $\dot{\overline{H}}=\dot{\bm q}^{\mathsf{T}}\bm\tau_e+\underbrace{R}_{\leq0}-\underbrace{\dot{\overline{\tilde{\bm q}}}^{\mathsf{T}}\bm S_2^{\mathsf{T}}\bigl(k_g\bm{sgn}(\dot{\overline{\tilde{\bm q}}})-\tilde{\bm\tau}_f\bigr)}_{\ge0} - \underbrace{\overline{\bm y}^{\mathsf{T}} \bm P\overline{\bm y}}_{\ge 0}\leq\dot{\bm q}^{\mathsf{T}}\bm\tau_e$      \\
0 & 1 &$-\dot{\bm q}^{\mathsf{T}}\underbrace{\overline{\bm F}_f}_{=\alpha_f\bm F_f}+\alpha_f\dot{\bm q}^{\mathsf{T}}{\bm F}_f+\underbrace{\dot{\overline{\hat{\bm q}}}^{\mathsf{T}}(\overline{\bm F}_f-\bm\tau_e)}_{=\dot{{\hat{\bm q}}}^{\mathsf{T}}(\overline{\bm F}_f-\bm\tau_e)<0}-\underbrace{\beta_o}_{\in\{0,1\}}\underbrace{\dot{{\hat{\bm q}}}^{\mathsf{T}}(\overline{\bm F}_f-\bm\tau_e)}_{<0}\leq0$&           \\
1 & 0 &      $-\underbrace{\dot{\bm q}^{\mathsf{T}}\overline{\bm F}_f}_{=\dot{\bm q}^{\mathsf{T}}\bm F_f>0}+\underbrace{\beta_f}_{\in\{0,1\}}\underbrace{\dot{\bm q}^{\mathsf{T}}{\bm F}_f}_{>0}+\underbrace{\dot{\overline{\hat{\bm q}}}^{\mathsf{T}}}_{=\alpha_o\dot{{\hat{\bm q}}}^{\mathsf{T}}}(\overline{\bm F}_f-\bm\tau_e)-\alpha_o\dot{{\hat{\bm q}}}^{\mathsf{T}}(\overline{\bm F}_f-\bm\tau_e)\leq 0$           &           \\
1 & 1 &       $-\underbrace{\dot{\bm q}^{\mathsf{T}}\overline{\bm F}_f}_{=\dot{\bm q}^{\mathsf{T}}\bm F_f>0}+\underbrace{\beta_f}_{\in\{0,1\}}\underbrace{\dot{\bm q}^{\mathsf{T}}{\bm F}_f}_{>0}+\underbrace{\dot{\overline{\hat{\bm q}}}^{\mathsf{T}}(\overline{\bm F}_f-\bm\tau_e)}_{=\dot{{\hat{\bm q}}}^{\mathsf{T}}(\overline{\bm F}_f-\bm\tau_e)<0}-\underbrace{\beta_o}_{\in\{0,1\}}\underbrace{\dot{{\hat{\bm q}}}^{\mathsf{T}}(\overline{\bm F}_f-\bm\tau_e)}_{<0}\leq0$         &        
\end{tblr}
\label{tab:passiv}
\end{table*}

Since the injected power is defined as $\dot{\bm q}^{\mathsf{T}}\bm\tau_e$, (\ref{equ:Hdot}) can be further rewritten as
\begin{align}
    \dot H=&\dot{\bm q}^{\mathsf{T}}\bm\tau_e-\dot{\bm q}^{\mathsf{T}}\bm F_f+\dot{\hat{\bm q}}^{\mathsf{T}}(\bm F_f-\bm\tau_e)\notag\\
    &-\dot{\tilde{\bm q}}^{\mathsf{T}}\bm S_2^{\mathsf{T}}\bigl(k_g\bm{sgn}(\dot{\tilde{\bm q}})-\tilde{\bm\tau}_f\bigr) - \bm y^{\mathsf{T}} \bm P\bm y.
    \label{equ:Hdot_pre}
\end{align}
Since the signs of $\dot{\bm q}^{\mathsf{T}}\bm F_f$ and $\dot{\hat{\bm q}}^{\mathsf{T}}(\bm F_f-\bm\tau_e)$ are unknown, the passivity of the system with respect to $\dot{\bm q}^{\mathsf{T}}\bm\tau_e$ cannot be guaranteed.
To ensure the passivity of the proposed controller, the following two virtual energy tanks are introduced~\cite{schindlbeck2015unified,shahriari2018valve}.\\
\noindent{\em - Torque Tank}:
To resolve the passivity violation due to the energy injected by the port $(\dot{\bm q},-\bm F_f)$, the first virtual energy tank is introduced, with its energy defined as $H_f=\frac{1}{2}t_f^2$. The dynamics of this energy tank is given by
\begin{align}
    \dot{t}_{f}&=\frac{\beta_f}{t_f}\gamma_f\dot{\bm q}^{\mathsf{T}}\bm F_f+\frac{\alpha_f}{t_f}(1-\gamma_f)\dot{\bm q}^{\mathsf{T}}\bm F_f,\\
\gamma_f&=\begin{cases}
1, & \text{if } \dot{\bm q}^{\mathsf{T}}\bm F_f > 0, \\[6pt]
0, & \text{else.}
\end{cases}
\end{align}
where $\gamma_f$ is a binary variable indicating whether the torque controller is passive, and $\beta_f$ and $\alpha_f$ are parameters that govern the capacity of the tank and will be defined later.
By employing this torque tank, the torque controller can be corrected as
\begin{align}
    \overline{\bm F}_f=\bigl(\gamma_f + \alpha_f(1-\gamma_f)\bigr)\bm F_f.
    \label{equ:F_cor}
\end{align}
The corrected torque controller will deactivate the torque-tracking task if the empty torque tank cannot absorb the unexpected power input from $(\dot{\bm q},-\bm F_f)$, further enhancing the passivity of the interactive system.\\
\noindent{\em - Observer Tank}:
When the corrected torque controller (\ref{equ:F_cor}) is applied, the input power port $(\dot{\hat{\bm q}},\overline{\bm F}_f-\bm\tau_e)$ remains of unknown sign. A second virtual energy tank, with its energy defined as $H_o=\frac{1}{2}t_o^2$, is introduced to enhance the passivity. The corresponding dynamics is formulated as
\begin{align}
    \dot{t}_{o}&=-\frac{\beta_o}{t_o}\gamma_o\dot{\hat{\bm q}}^{\mathsf{T}}(\overline{\bm F}_f-\bm\tau_e)+\frac{\alpha_o}{t_o}(\gamma_o-1)\dot{\hat{\bm q}}^{\mathsf{T}}(\overline{\bm F}_f-\bm\tau_e),\\
\gamma_o&=\begin{cases}
1, & \text{if } \dot{\hat{\bm q}}^{\mathsf{T}}(\overline{\bm F}_f-\bm\tau_e) < 0, \\[6pt]
0, & \text{else.}
\end{cases}
\end{align}
where $\gamma_o$ is a binary variable indicating whether the observer is passive.
Along with $\beta_f,\alpha_f$, the parameters $\beta_o,\alpha_o$ are related to the tank capacity and are defined as follows:
\begin{align}
    \beta_{f,o} &=
\begin{cases}
1, & \text{if } H_{f,o} \le U_{f,o}, \\[6pt]
0, & \text{else.}
\end{cases}\\
\alpha_{f,o}& =
\begin{cases}
1, & \text{if } H_{f,o} \ge L_{f,o} + \delta_{f,o}, \\[1pt]
\dfrac{1}{2}\left[1 - \cos\left(\dfrac{H_{f,o} - L_{f,o}}{\delta_{f,o}} \pi\right)\right],
& \text{if } H_{f,o} \leq L_{f,o} + \delta_{f,o} \\
&\text{and }H_{f,o}\ge L_{f,o}, \\[1pt]
0, & \text{else.}
\end{cases}
\end{align}
where $U_{f,o},L_{f,o}$ are the upper and lower limits of storage of the torque and observer tank, and $\delta_{f,o}$ governs the transition process as the tanks approach empty.
The corrected observer is described by 
\begin{align}
    \dot{\overline{\hat{\bm q}}}=\bigl(\gamma_o + \alpha_o(1-\gamma_o)\bigr)\dot{\hat{\bm q}}.
    \label{equ:q_cor}
\end{align}
It can be proven that with the corrected torque controller (\ref{equ:F_cor}) and observer (\ref{equ:q_cor}), the passivity of the system with respect to $\dot{\bm q}^{\mathsf{T}}\bm\tau_e$ is guaranteed.\\
\noindent\textbf{Stability Analysis}:
With the passivity-based correction, the storage function for the whole system is formulated as
\begin{align}
    \overline{H}=&\frac{1}{2}\dot{\overline{\tilde{\bm q}}}^{\mathsf{T}}\bm M(\bm q)\dot{\overline{\tilde{\bm q}}}+\frac{1}{2}\bm z^{\mathsf{T}}\bm B\bm z+\frac{1}{2}\bm s^{\mathsf{T}}\bm B\bm s\notag\\
    &+\frac{1}{2}\Delta\hat{\bm\theta}^{\mathsf{T}}(2\alpha_{\theta}\bm K_s+\bm K_{\theta})\Delta\hat{\bm\theta}+H_f+H_o,
    \label{equ:wholeH}
\end{align}
where $\dot{\overline{\tilde{\bm q}}}\triangleq\dot{\bm q}-\dot{\overline{\hat{\bm q}}}$. Introducing $\overline{\bm y}=[\dot{\overline{\tilde{\bm q}}}^{\mathsf{T}},\Delta{\hat{\bm\theta}}^{\mathsf{T}} ,\Delta\dot{\hat{\bm\theta}}^{\mathsf{T}},\bm e^{\mathsf{T}}]^{\mathsf{T}}$, the differential of the storage function is 
\begin{align}
    \dot{\overline{H}}=&\dot{\bm q}^{\mathsf{T}}\bm\tau_e-\dot{\bm q}^{\mathsf{T}}\overline{\bm F}_f+\bigl(\beta_f\gamma_f+\alpha_f(1-\gamma_f)\bigr)\dot{\bm q}^{\mathsf{T}}\bm F_f\notag\\
    &+\dot{\overline{\hat{\bm q}}}^{\mathsf{T}}(\overline{\bm F}_f-\bm\tau_e)-\bigl(\beta_o\gamma_o+\alpha_o(1-\gamma_o)\bigr)\dot{\hat{\bm q}}^{\mathsf{T}}(\overline{\bm F}_f-\bm\tau_e)\notag\\
    &-\dot{\overline{\tilde{\bm q}}}^{\mathsf{T}}\bm S_2^{\mathsf{T}}\bigl(k_g\bm{sgn}(\dot{\overline{\tilde{\bm q}}})-\tilde{\bm\tau}_f\bigr) - \overline{\bm y}^{\mathsf{T}} \bm P\overline{\bm y}.
    \label{equ:Hdot_final}
\end{align}
Under different configurations and with the definition of the tank-related terms denoted as $R$, the passivity of the whole system is guaranteed, as shown in Table~\ref{tab:passiv}.

For the assistance delivery task, we can now state the following Theorem.\\
\noindent\textbf{Theorem 1}:
{\em The proposed controller (\ref{equ:desire_th}), (\ref{equ:u_direct}), and (\ref{equ:u_sea}) with passivity-based corrections (\ref{equ:F_cor}) and (\ref{equ:q_cor}) guarantees the performance of the torque-tracking task if the passivity is not violated.}
\begin{IEEEproof}
Considering the interactive environment and the global stable state, the following Lyapunov-like candidate is proposed:
\begin{align}
    V=\overline{H}+H_{env}(\bm x_{env})-\frac{1}{2}t_f^{'2}-\frac{1}{2}t_o^{'2},
    \label{equ:Lya_all}
\end{align}
where $H_{env}(\bm x_{env})\in \mathbb{R}_{\ge 0}$ is the non-negative energy function of the environment, with $\bm x_{env}$ representing the deviation from the stable environment state, while $t_f^{'2},t_o^{'2}$ are stable tank states that indicate the minimum system energy.

When $k_g>\zeta$ and (\ref{equ:con1})--(\ref{equ:con3}) are satisfied, 
differentiating (\ref{equ:Lya_all}) with respect to time and substituting (\ref{equ:Hdot_final}) into it yields the following expression:
\begin{align}
    \dot{V}=&\dot{\bm q}^{\mathsf{T}}\bm\tau_e+\dot{H}_{env}(\bm x_{env})+\underbrace{R}_{\leq0}\notag\\
    &-\underbrace{\dot{\overline{\tilde{\bm q}}}^{\mathsf{T}}\bm S_2^{\mathsf{T}}\bigl(k_g\bm{sgn}(\dot{\overline{\tilde{\bm q}}})-\tilde{\bm\tau}_f\bigr)}_{\ge0} - \underbrace{\overline{\bm y}^{\mathsf{T}} \bm P\overline{\bm y}}_{\ge 0}.
\end{align}

Dissipation can always be presumed in the environment, and the following inequalities always hold~\cite{ott2008cartesian,haddadin2024unified}:
\begin{align}
    \dot{H}_{env}(\bm x_{env})\leq&-\dot{\bm q}^{\mathsf{T}}\bm\tau_e.
\end{align}

Thus, $\dot{V}<0$ can be derived. As $V>0$, the overall system is asymptotically stable, indicating the convergence of $\dot{\overline{\tilde{\bm q}}}\rightarrow0$ and $\Delta\bm\theta\rightarrow0$.
Under the condition that passivity is not violated, namely $\gamma_f=\gamma_o=1$, the original design holds: $\overline{\bm F}_f=\bm F_f, \dot{\overline{\hat{\bm q}}}=
\dot{\hat{\bm q}}$.
According to (\ref{equ:dynamic2}) and Proposition 1, the performance of the torque-tracking task is guaranteed.
\end{IEEEproof}

\section{Experiments}
\begin{figure}[!t]
    \centering
    \includegraphics[width=0.7\linewidth]{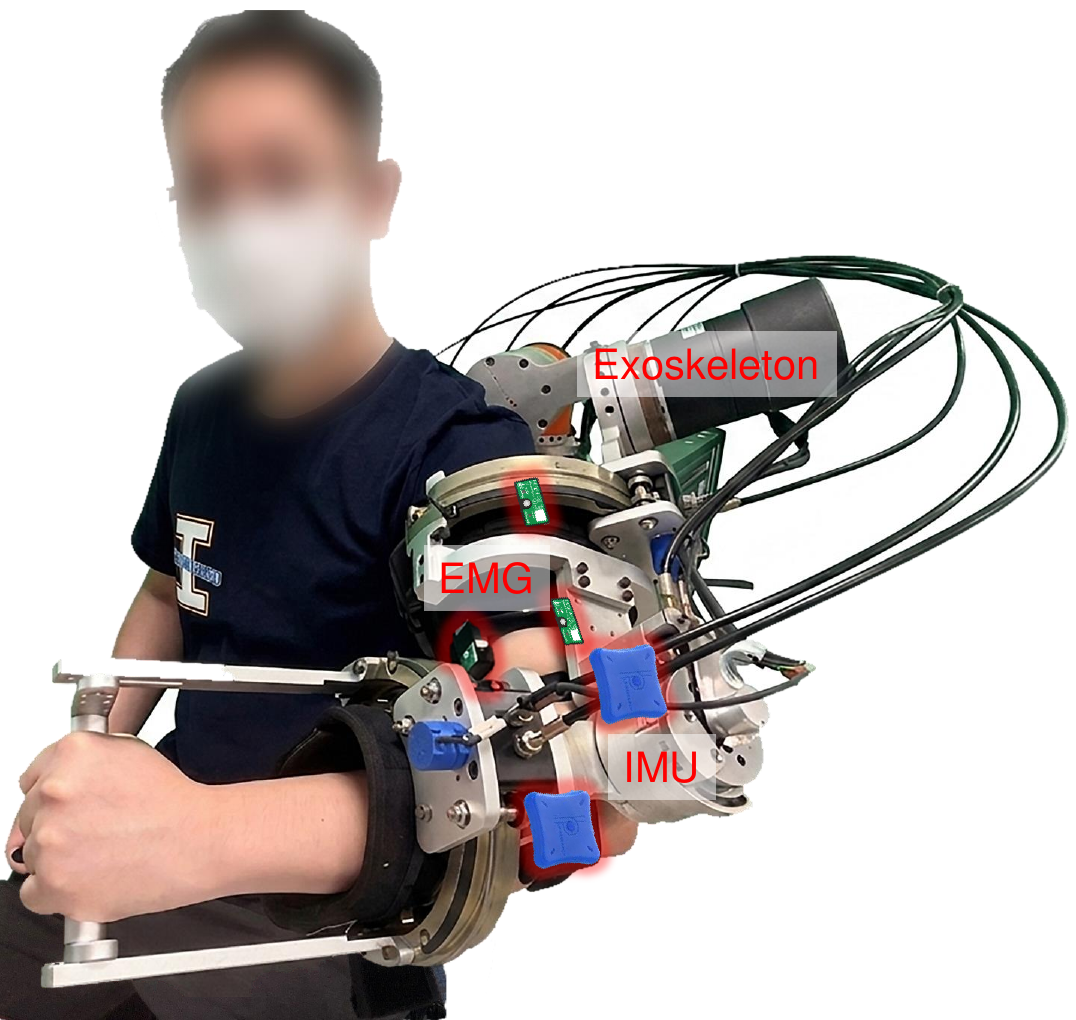}
    \caption{\textbf{Experimental setup of the upper-limb exoskeleton.} IMUs are worn by the subject to measure limb angles for estimating the anatomical torque reference, while EMG sensors are mounted on different upper-limb muscles to evaluate region-specific muscle activity.}
    \label{equipment}
    \vspace{-0.3cm}
\end{figure}

The proposed control framework was implemented on a custom-developed upper-limb exoskeleton, as shown in Fig.~\ref{equipment}. The two direct-drive shoulder joints, namely Joint~1 and Joint~2, are actuated by direct-drive joint modules (RJSIIT-17-RevB2 and RJSIIT-17-RevB5).
Joints 3 to 5 employ harmonic-drive servo motors (AK80-64) on the motor side, which are connected to elastic elements through steel cables to form cable-driven SEAs. On the joint side, encoders are installed to measure joint motion, including 3590S-2-104L encoders for the internal/external rotation joints and a QY2204-SSI encoder for the elbow joint. Torque sensors (TK17-191151) are also integrated to measure interaction information between the wearer and the exoskeleton.
The exoskeleton arm weighs 7.78~$\mathrm{kg}$, which helps reduce inertial effects on the wearer. It can provide more than $48~\mathrm{N\cdot m}$ of assistance across all joints and achieves a backdrivability torque of less than $0.7~\mathrm{N\cdot m}$ at a joint speed of 10$\mathrm{^{\circ}/s}$. The exoskeleton is controlled by a microcontroller unit operating at a control frequency of $1~\mathrm{kHz}$, which sends commands to the motor drivers running at $4~\mathrm{kHz}$.
During the experiments, the subjects were required to wear inertial measurement units (IMUs) operating at $200~\mathrm{Hz}$ to measure limb angles, which were then used to drive the musculoskeletal system. In addition, three wireless surface electromyography (EMG) sensors (Ws450, Biometrics Ltd.) were mounted on the human arm, specifically on the biceps brachii (BB), triceps brachii (TB), and deltoid (DELT) muscles, to evaluate the controller’s ability to reduce muscle loading in different parts of the upper limb.

The interactive control algorithm was programmed and implemented on the microcontroller unit, whereas the torque generation module was executed on an additional PC equipped with an Intel i5-13490F CPU and an RTX 4060Ti GPU. The two systems communicated through a CAN bus at a frequency of $1~\mathrm{kHz}$, ensuring real-time execution of computationally intensive modules, such as musculoskeletal simulation and online torque refinement. All PC-side modules were integrated within Robot Operating System 2 (ROS~2)~\cite{Macenski2022RobotOS}.
The dynamic parameters of the exoskeleton were obtained using the open-source Orocos Kinematics and Dynamics Library\footnotemark, enabling model-based interactive control.\footnotetext{\url{https://www.orocos.org/wiki/orocos/kdl-wiki.html}}

\begin{table}[t]
\centering
\caption{Hyperparameters for muscle controller training}
\label{table:rl_params}
\begin{tblr}{
  colspec = {l l r},
  column{1} = {font=\bfseries},
  row{1}    = {font=\bfseries},
  cell{2}{1} = {r=3}{},
  cell{5}{1} = {r=5}{},
  cell{10}{1} = {r=14}{},
  hline{1,24} = {-}{0.08em},
  hline{2,5,10} = {-}{0.03em},
  rowsep = 1.5pt,
}
Category             & Parameter                          & Value                     \\
Network Architecture & Actor                              & $512 \to 256 \to 128$     \\
                     & Critic                             & $512 \to 256 \to 128$     \\
                     & Activation                         & ELU                       \\
Reward Weights       & Initial $\sigma_p$                         & $0.3$                       \\
                     & Initial $\sigma_{\dot{p}}$                 & $30$                       \\
                     & $\lambda_a$                        & $1.5$                       \\
                     & $\alpha_p$                         & $0.001$                       \\
                     & $\alpha_{\dot{p}}$                 & $0.001$                       \\
PPO Algorithm        & Learning rate                      & $3\times10^{-4}$          \\
                     & Learning rate schedule              & Adaptive                  \\
                     & Rollout steps per env.             & $24$                      \\
                     & Number of minibatches              & $4$                       \\
                     & Discount factor $\gamma$           & $0.99$                    \\
                     & GAE parameter $\lambda$            & $0.95$                    \\
                     & Policy clip                        & $0.2$                     \\
                     & Value clip                         & $0.2$                     \\
                     & Max. gradient norm                  & $1.0$                     \\
                     & Update epochs       & $5$                       \\
                     & Initial standard deviation              & $1.0$                     \\
                     & Learnable standard deviation                      & Yes                       \\
                     & Entropy coefficient                & $0.001$                   \\
                     & Value coefficient                  & $1.0$                     \\
\end{tblr}
\vspace{-0.3cm}
\end{table}


 We evaluated our framework through the following experiments.
\begin{enumerate}
\item [-]\textit{Muscle Controller}: 
This experiment was designed to validate the proposed muscle controller. The controller was trained using parallel simulation environments, and its efficiency and convergence behavior were evaluated. To assess the benefit of jointly training the controller on multiple trajectories, we compared a controller trained on a large-scale trajectory dataset with controllers trained separately on individual motion segments. The generalization capability of the proposed controller was further examined in a real-world setup, in which target trajectories were obtained from the free upper-limb movements of a subject wearing IMUs.
\item [-]\textit{Passivity-based Controller}: 
To validate the proposed passivity-based interaction torque controller, we developed a single-joint SEA simulation model and conducted a series of numerical simulations. The stability and passivity-preservation capability of the controller were examined in scenarios where controller behavior switching was triggered by periodic activation of the energy tank. In addition, real-world experiments were performed to demonstrate torque-tracking performance and control-priority switching under different human--robot interaction conditions, using both a direct-drive shoulder joint and a cable-driven elbow joint.
\item [-]\textit{Anatomical Assistance}: 
The proposed ATP controller was implemented on the cable-driven upper-limb exoskeleton to validate its effectiveness in delivering anatomical assistance. Ablation studies were conducted to evaluate the contribution of anomaly-score guidance to torque refinement under two anomalous conditions, specifically simulated collisions and motion near kinematic singularities.
An additional ablation experiment was performed to demonstrate the effectiveness of the torque-refinement module. Furthermore, a group of participants was recruited to compare the proposed ATP controller with several baseline methods. EMG signals from the BB, TB, and DELT were used as evaluation metrics to quantify the effects of ATP on different muscles during various upper-limb movements.
\end{enumerate}

\begin{table*}
\centering
\caption{Performance Comparison across Datasets}
\label{tab:performance}
\setlength{\tabcolsep}{3pt} 
\begin{tblr}{
  colspec={X[1.2,l]X[0.6,r]|X[0.5,r]X[0.5,r]X[0.5,r]X[0.5,r]|X[0.5,r]X[0.5,r]X[0.5,r]X[0.5,r]|X[0.8,c]},
  cell{1}{1}={r=2}{l},
  cell{1}{3}={c=4}{c},
  cell{1}{7}={c=4}{c},
  cell{1}{11}={r=2}{c},
  row{1-2}={font=\bfseries\small},
  hline{1,3,8}={0.08em},
  hline{2}={3-10}{},
  cell{6,7}{3-11} = {bg=gray!30},
}
Dataset & Length & \SetCell[c=4]{c} Jointly Trained Policy & & & & \SetCell[c=4]{c} Separately Trained Policy & & & & Ratio \\
        & (min)  & Sh1 & Sh2 & Sh3 & Elb & Sh1 & Sh2 & Sh3 & Elb & \\
ACCAD       & 26.7 & 0.170(4) & 0.164(3) & 0.157(2) & 0.139(6) & 0.112(13) & 0.137(13) & 0.079(7) & 0.080(7) & 1.55(64.5$\%$) \\
HumanEva    & 8.5  & 0.150(2) &  0.145(1)  &  0.106(1) & 0.148(1) & 0.100(5) &  0.168(11) & 0.067(1) & 0.127(14) & 1.19(84.1$\%$) \\
SSM\_synced & 1.9  & 0.196(5) &  0.464(197) &  0.203(2) &  0.367(26)
 & 0.121(10) &  0.430(108) &  0.117(2)  & 0.182(16) & 1.45(69.0$\%$) \\
Lafan1      & 3.8  & 0.220 & 0.208 & 0.236 & 0.180 & 0.259 & 0.393 & 0.252 & 0.127 & 0.82(122.1$\%$) \\
Ours        & 4.8  & 0.038    & 0.061    & 0.025    & 0.019    & 0.030 & 0.195 & 0.035 & 0.020 & 0.51(194.1$\%$) \\
\end{tblr}
\begin{tablenotes}
\footnotesize
\item Note: Performance is evaluated by RMSE in radians. Ratio denotes the performance ratio of the Jointly Trained Policy relative to the Separately Trained Policy, with a lower value indicating better relative performance and a higher level of performance retention. Sh: Shoulder; Elb: Elbow. The values in parentheses represent the standard deviation. For example, 0.170 (4) means 0.170 ± 0.004.
\end{tablenotes}
\end{table*}

The experimental protocol was approved by the ethics committee of XXX in December 2024. All participants provided written informed consent before taking part in the experimental sessions.




\begin{figure}[!t]
    \centering
    \vspace{-0.65cm}
    \includegraphics[width=0.9\linewidth]{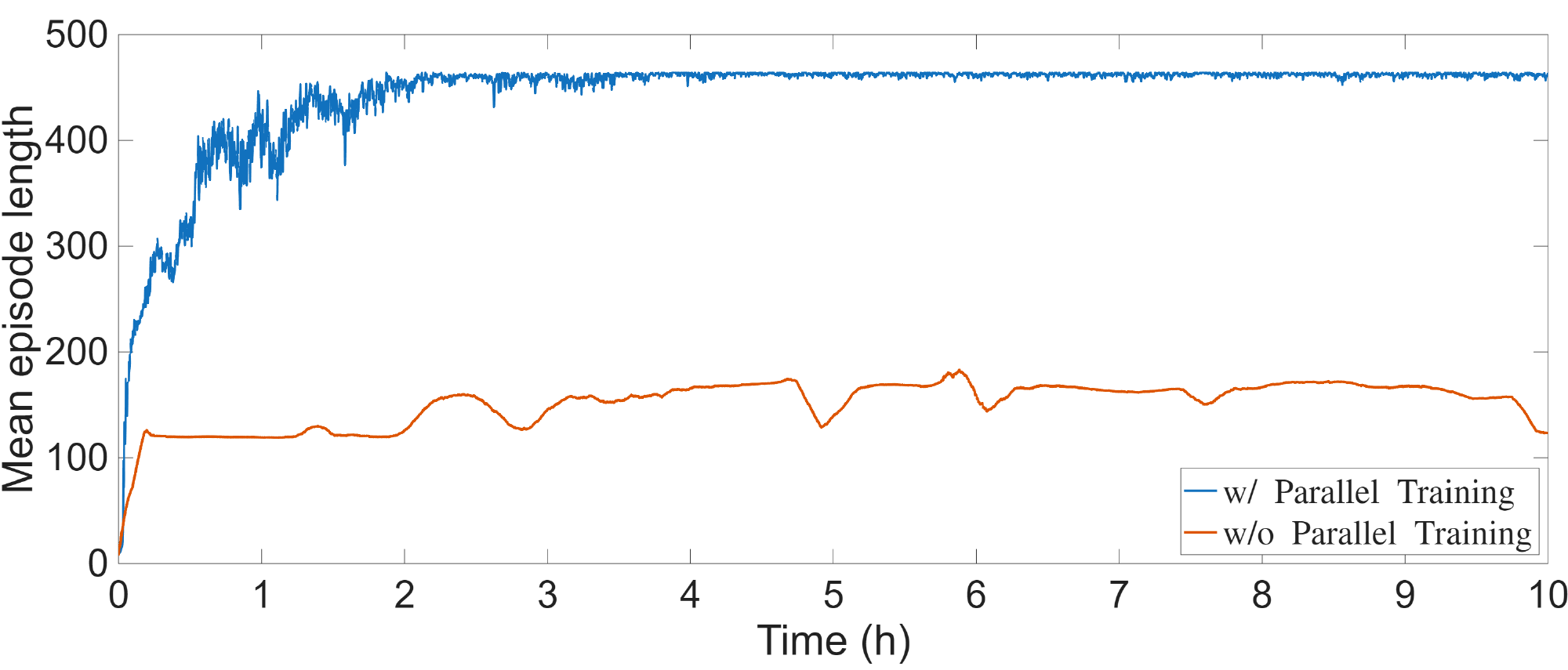}
    \caption{\textbf{Comparison of training efficiency.} The performances are evaluated based on the mean episode length during training with and without parallelization.}
    \label{fig:pickup}
    \vspace{-0.65cm}
\end{figure}

\section{Results}
\subsection{Muscle Controller}
To obtain the anatomical torque reference for the upper-limb exoskeleton, the muscle controller, which accounts for high-fidelity muscle behavior, generates joint torques for motion tracking. The muscle controller was trained in a parallel manner using MyoArm in a GPU-accelerated environment, with the relevant training parameters outlined in Table~\ref{table:rl_params}.
We compared the training results between a single-environment setup and a parallel training environment, using the arm raising task to validate the improvements offered by GPU-accelerated training. The number of environments was set to 256. As shown in Fig.~\ref{fig:pickup}, without parallel training, the policy struggled to complete the trajectory and ultimately failed due to tracking errors exceeding the boundary. In contrast, the policy trained in multiple environments successfully completed the task after 3 hours of training. Although the motion imitation task was simplified to focus on a basic arm raising movement, the inherent nonlinearity of the musculoskeletal system and the high-dimensional action space necessitated at least 3 hours to achieve stable training performance. For more complex upper-limb dynamic tasks and long-duration reference trajectories, parallel training becomes even more essential to improve training efficiency.

\begin{figure}[!t]
    \centering
    \vspace{-0.65cm}
    \includegraphics[width=0.9\linewidth]{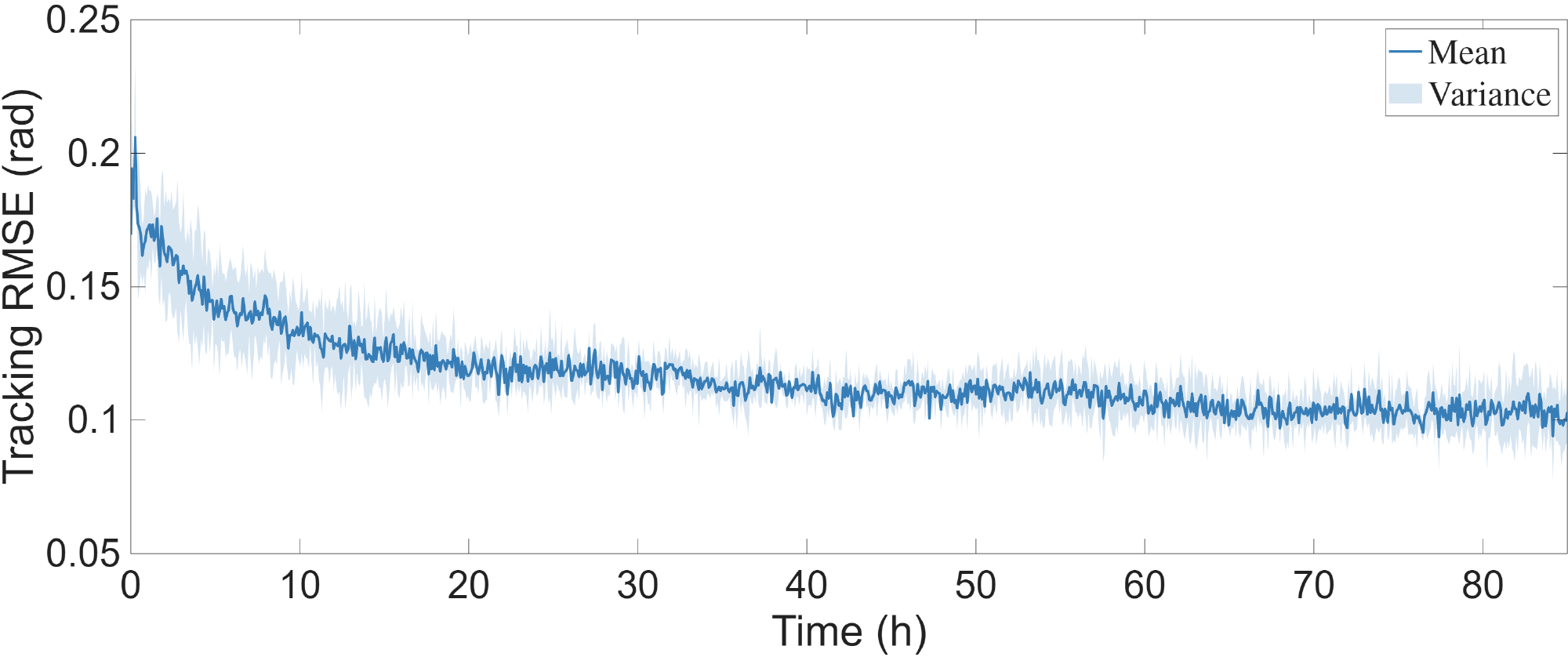}
    \caption{\textbf{Training curve of the muscle controller.} The tracking error across different tasks decreases as the controller continues to be jointly trained on all datasets.}
    \label{fig:simu_train}
    \vspace{-0.3cm}
\end{figure}

\begin{figure*}[!t]
    \vspace{-0.5cm}
    \centering
    \subfigure[ ]{
        \label{fig:simu1}
        \includegraphics[width=0.45\linewidth]{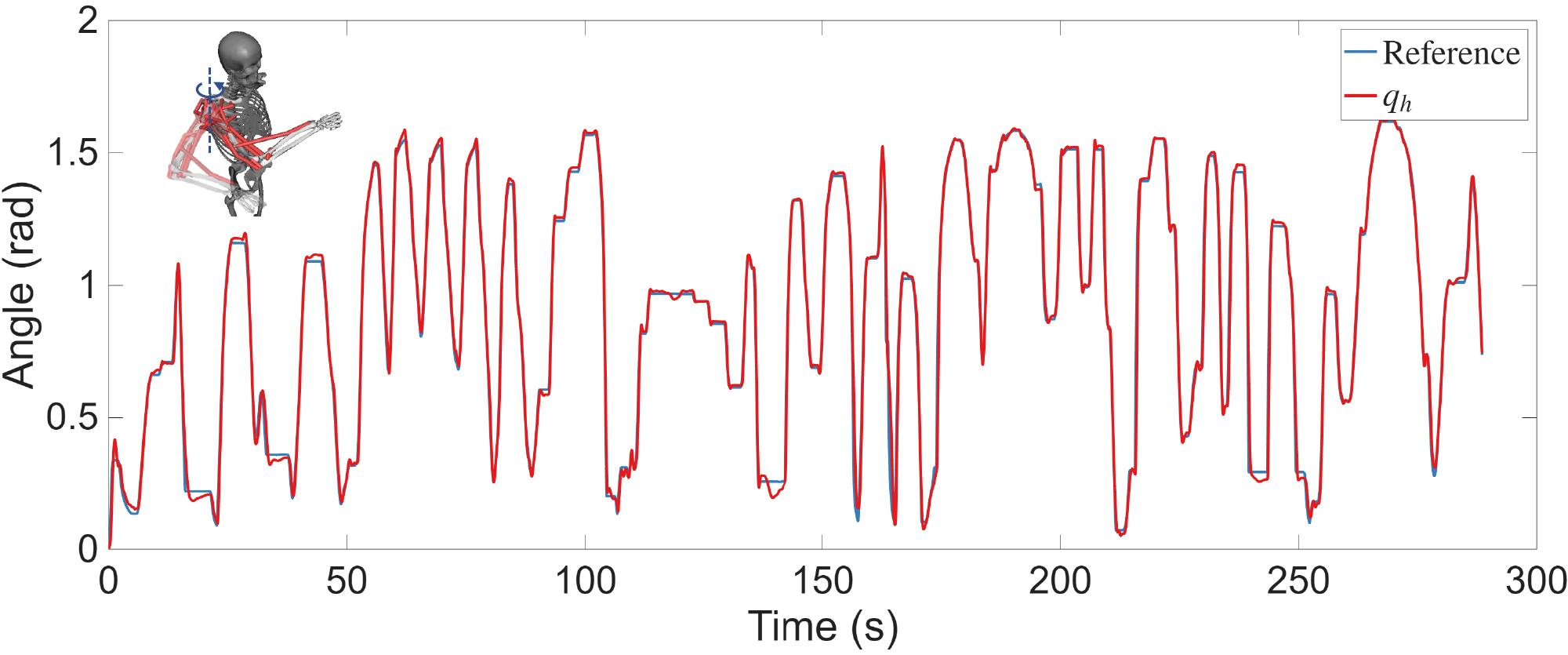}
    }\hfill
    \subfigure[ ]{
        \label{fig:simu2}
        \includegraphics[width=0.45\linewidth]{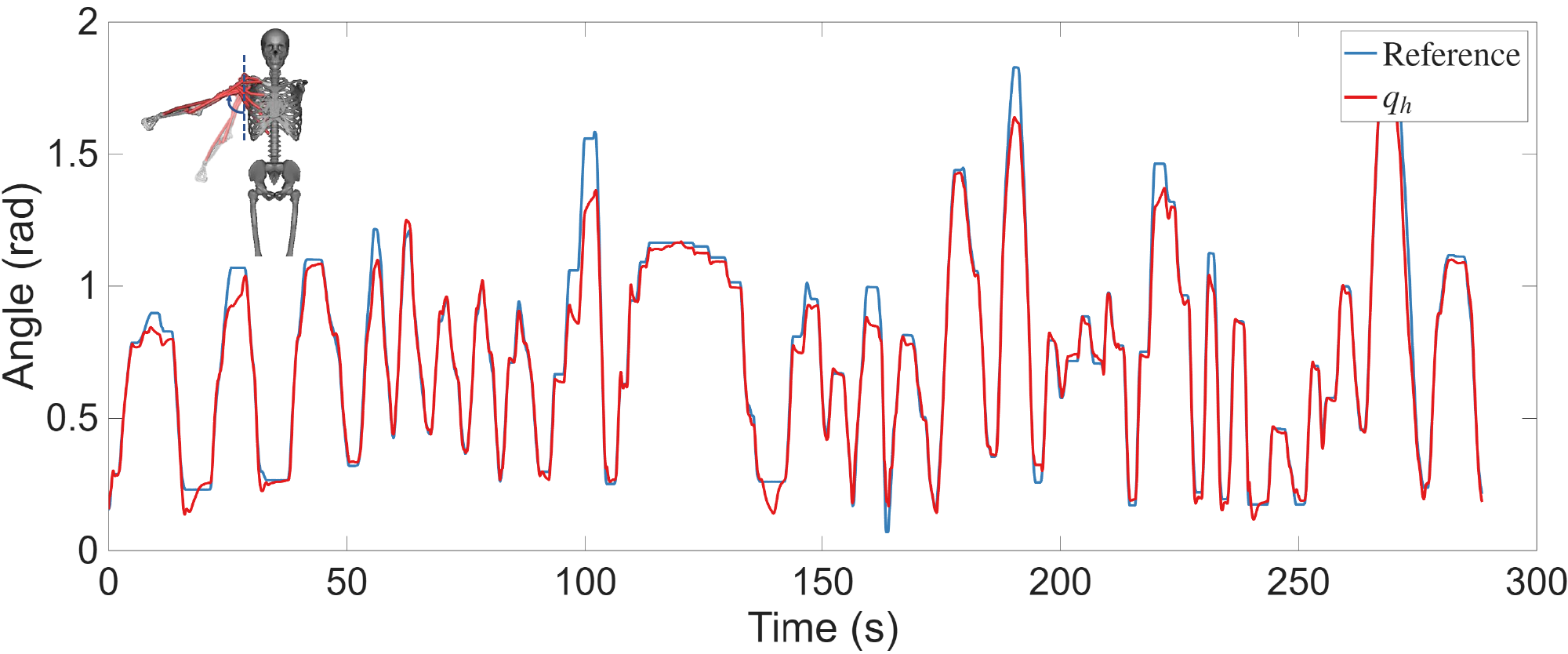}
    }
    \subfigure[ ]{
        \label{fig:simu3}
        \includegraphics[width=0.45\linewidth]{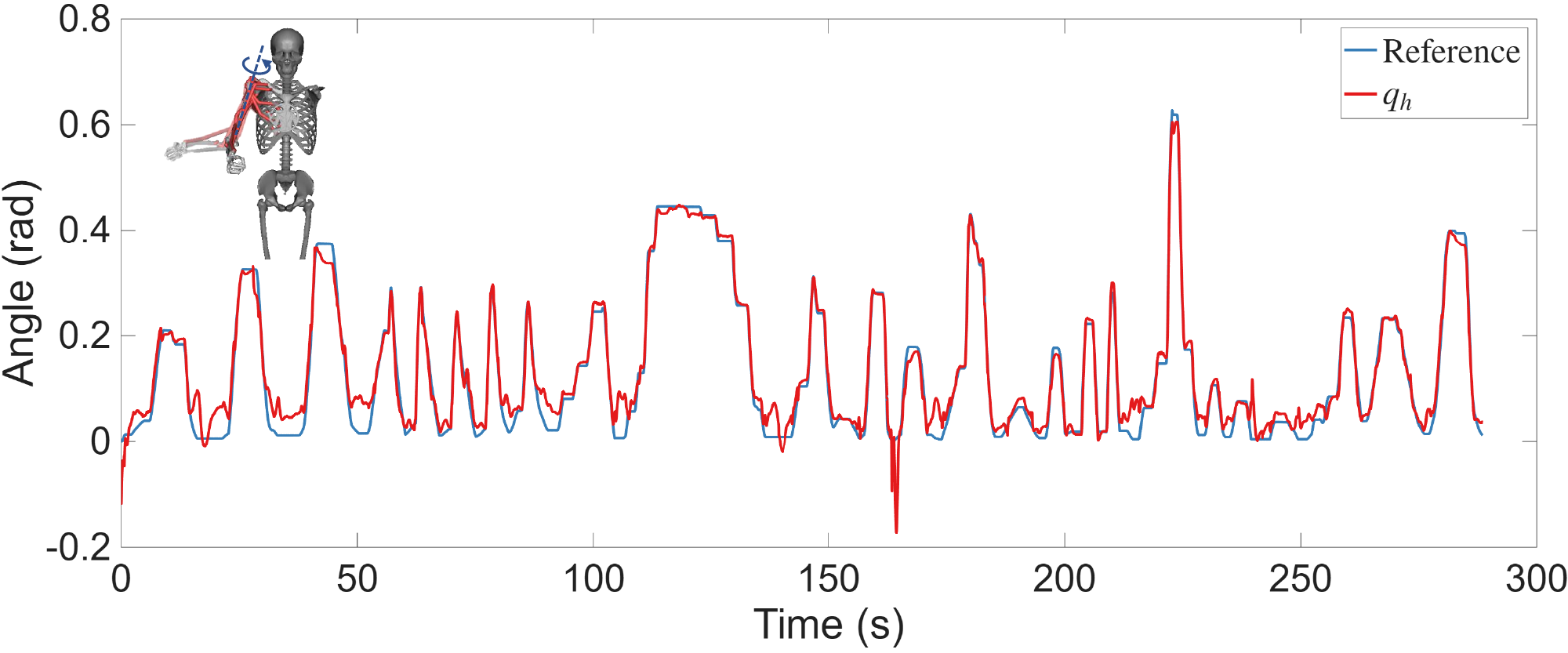}
    }\hfill
    \subfigure[ ]{
        \label{fig:simu4}
        \includegraphics[width=0.45\linewidth]{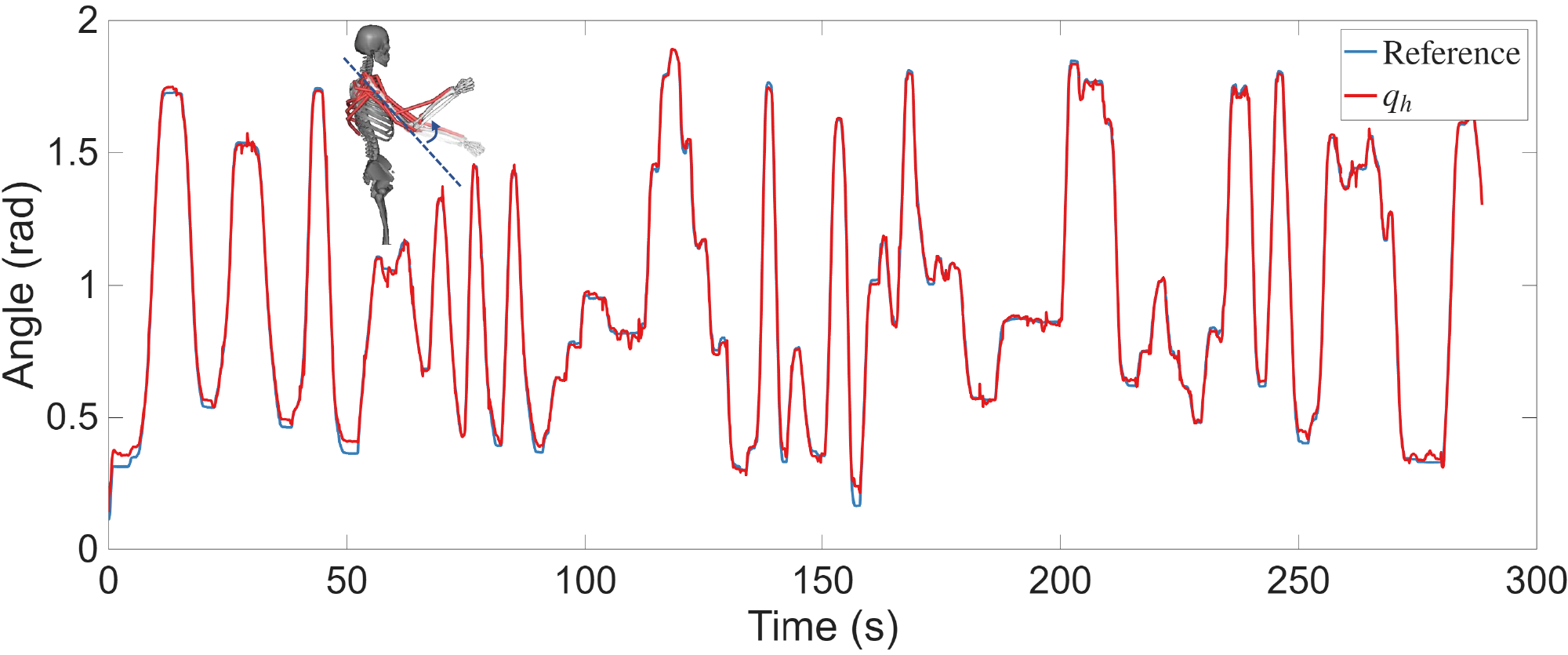}
    }
    \caption{\textbf{Motion-tracking performance of the musculoskeletal system driven by the muscle controller.} The controller is jointly trained on multiple trajectory datasets and then used to coordinate the activation of all muscles, enabling four independent joints to track a long-duration motion sequence collected from the exoskeleton.}
    \label{fig:simu_track}
    \vspace{-0.5cm}
\end{figure*}

To generate anatomical torque references that generalize across diverse upper-limb motions, we collected 30 motion sequences to train the unified muscle control policy. These sequences were selected from multiple datasets containing upper-limb movements. The adopted dataset consisted of 18 sequences from the ACCAD subset of the AMASS dataset, 5 sequences from the HumanEva dataset, 5 sequences from the SSM dataset, 1 long-duration dancing sequence from Lafan, and 1 long-duration sequence collected from the exoskeleton, in which the wearer freely moves the arm to cover a broad range of motion patterns relevant to exoskeleton assistance. The policy was jointly trained on all datasets, allowing it to learn multiple upper-limb movements simultaneously. The number of environments was set to 1024 to ensure training efficiency. 
The improvement achieved by the policy was quantified by the root mean square error (RMSE) at each epoch during training, as shown in Fig.~\ref{fig:simu_train}. The tracking error across the entire dataset decreased progressively during training and reached 0.1 $\mathrm{rad}$ as training converged.
Figure~\ref{fig:simu_track} illustrates the tracking performance of the three shoulder joints and the elbow joint on a long-duration motion sequence collected from the exoskeleton. The RMSE of tracking, averaged over four joints, is 0.036 $\mathrm{rad}$, which satisfies the requirement for accurate computation of anatomical torque references for exoskeleton assistance.
Although the muscle controller was trained on a large and diverse set of upper-limb movements, the inherent nonlinearity of the system can still cause unexpected over-movements during motion, suggesting that the controller may produce excessively large joint torques. Therefore, online torque refinement is necessary to further constrain the anatomical reference torque and ensure safe assistance.

We also compared the jointly trained muscle controller with the controllers that were separately trained on each sequence.
The results are summarized in Table~\ref{tab:performance}.
The jointly trained policy retained approximately $65$\%--$85\%$ of the performance achieved by policies trained on individual trajectories for short motion clips, while outperforming them on the long-duration tasks (Lafan1 and Ours).
This is mainly because the jointly trained policy is able to explore areas that the single-trajectory policy cannot reach, thus further improving trajectory-tracking performance. Additionally, aside from the tracking bias caused by rapid arm shaking in the SSM\_synced dataset, the jointly trained policy demonstrated more stable tracking performance with smaller variance than the separate-trained policy, indicating greater robustness.

\begin{figure}[!t]
    \centering
    \includegraphics[width=0.95\linewidth]{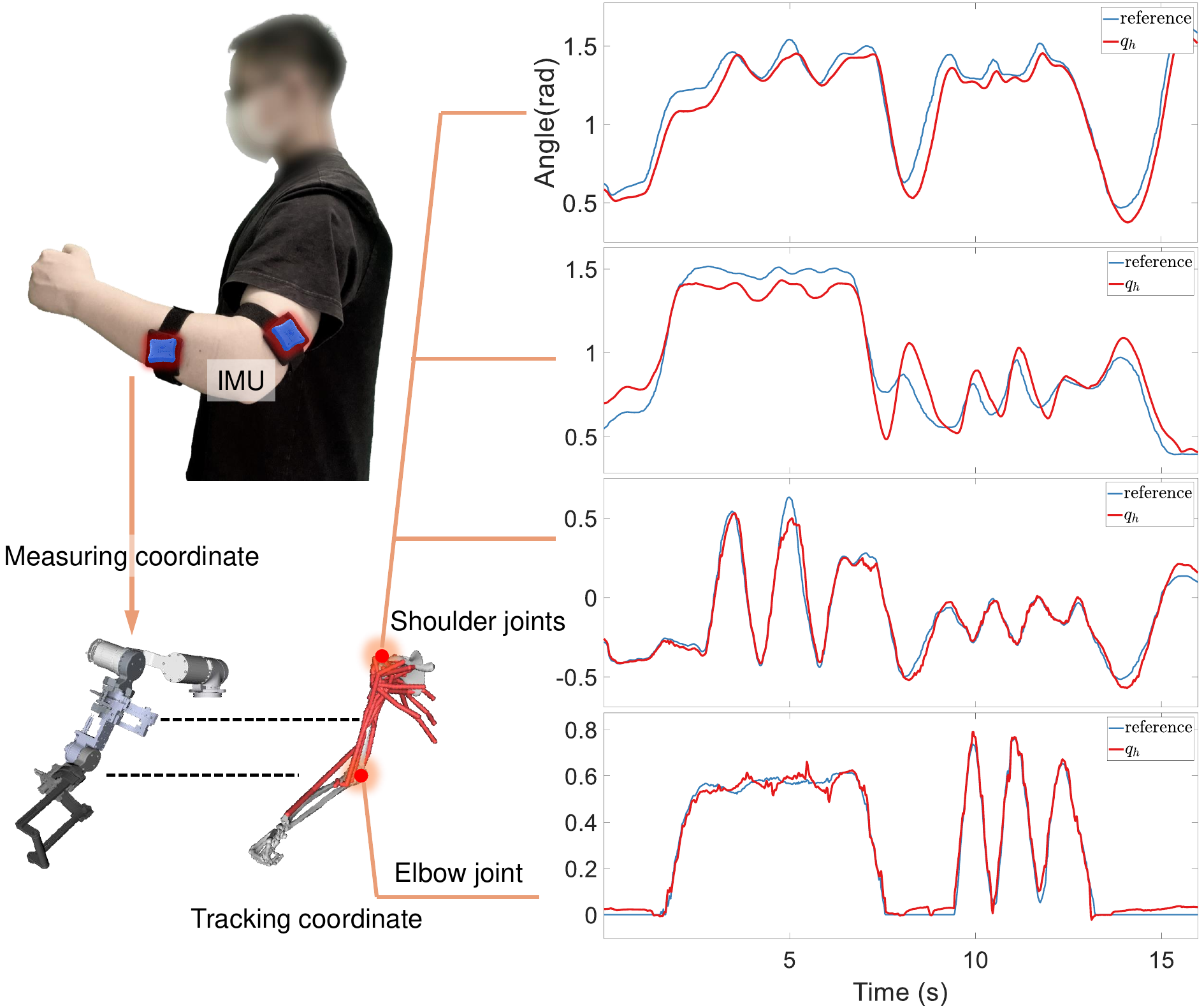}
    \caption{\textbf{Position tracking achieved by applying the muscle controller to unseen upper-limb movements in a real-world setup.} The target positions are obtained by measuring joint angles in the exoskeleton coordinate using IMUs and subsequently mapping them to the musculoskeletal coordinate. The muscle controller then drives three shoulder joints and one elbow joint to achieve real-time upper-limb pose tracking.}
    \label{fig:imu_tracking}
\end{figure}

To evaluate the generalization capability of the muscle controller during the real assistance phase, we conducted an experiment in which the controller was required to track position commands measured by IMUs. In this experiment, the subject wore the IMUs and performed free upper-limb movements. The corresponding joint angles were measured in the exoskeleton coordinate and mapped to the four joint angles of the musculoskeletal system, denoted as $\bm q_h$, which served as the tracking targets for the muscle controller. To ensure real-time tracking and provide accurate joint torque estimates for the subsequent exoskeleton assistance phase, the musculoskeletal system was run independently on the CPU in headless mode. This configuration increased the simulation speed by a factor of 20, allowing it to match the IMU sampling rate of $200~\mathrm{Hz}$ and enabling real-time controller responses during dynamic movements.
As shown in Fig.~\ref{fig:imu_tracking}, the tracking RMSEs for the three shoulder joints were 0.1 $\mathrm{rad}$, 0.107 $\mathrm{rad}$, and 0.048 $\mathrm{rad}$, respectively, while the RMSE for the elbow joint was 0.029 $\mathrm{rad}$. These results demonstrate that the proposed muscle controller can accurately track the wearer’s upper-limb motion and generalize to previously unseen movements. Based on the accurate kinematic alignment between the musculoskeletal system and the wearer, the internal joint torques generated by the model can be used as biological references for anatomical assistance.

\begin{figure}[!t]
    \centering
    \includegraphics[width=0.95\linewidth]{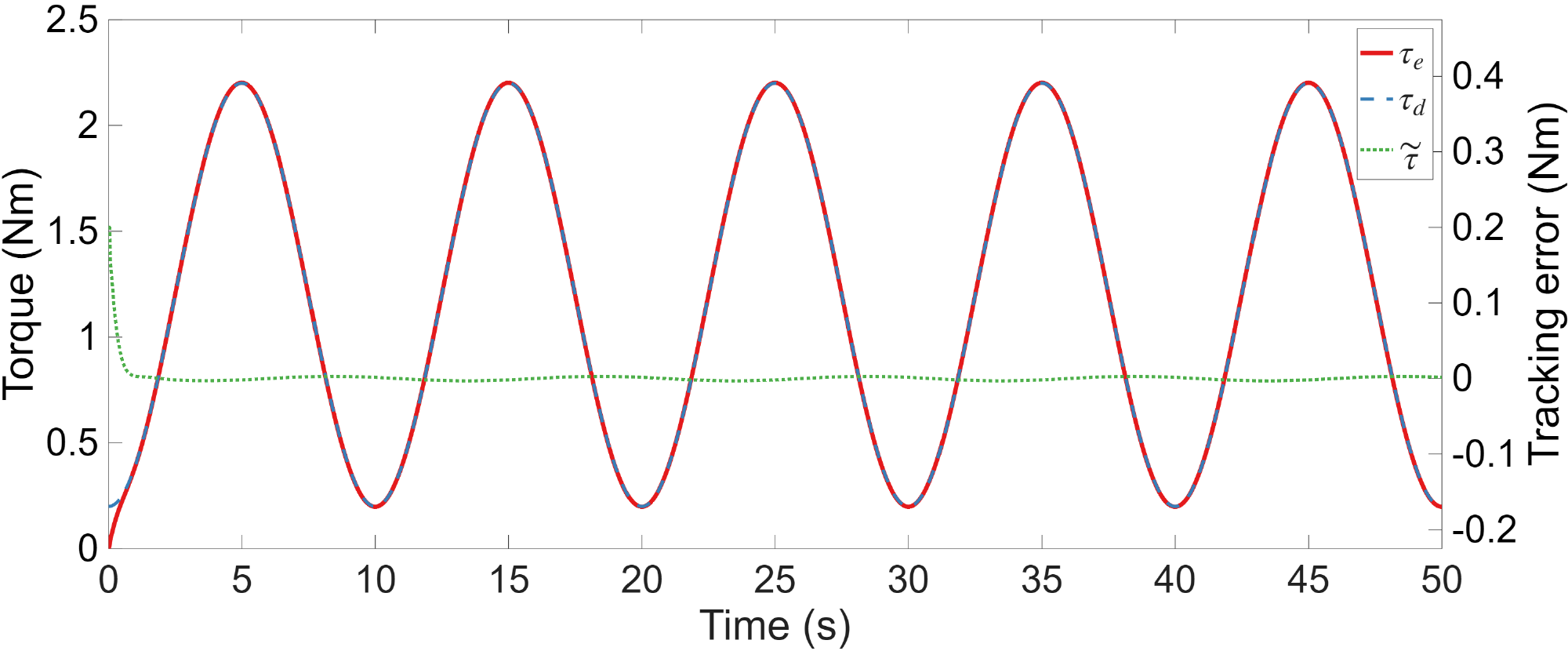}
    \caption{\textbf{Interaction torque-tracking performance in simulation.} The proposed passivity-based controller is implemented on a single-joint SEA in simulation and achieves accurate torque tracking when passivity is not violated.}
    \label{fig:good_force_tracking}
    \vspace{-0.65cm}
\end{figure}

\subsection{Passivity-based Controller}
We validated the proposed energy tank-based controller on a simulated single-joint SEA. To simulate human--robot interaction and external energy injection into the robotic system, the interaction was modeled as a moving contact surface with unilateral contact: contact forces were generated only when the joint motion pushed against the surface. This setup approximates the close physical coupling between the wearer and the exoskeleton during assistance. The contact surface was prescribed to follow a forced sinusoidal motion, thereby periodically injecting energy into the system to evaluate the passivity guarantee of the proposed controller.

\begin{figure}[!t]
    \centering
      \subfigure[ ]{
          \label{fig:tracking_f}
          \includegraphics[width=0.45\linewidth]{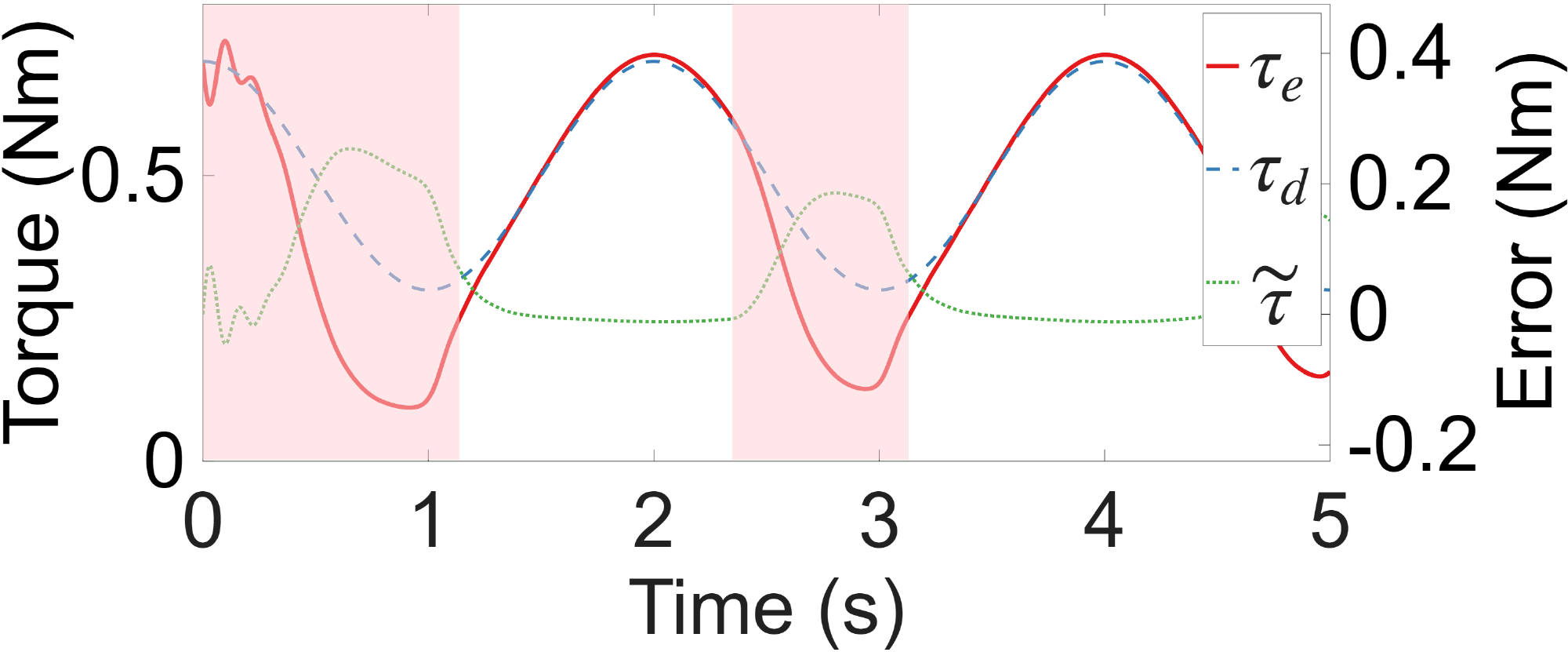}
          }\hfill
      \subfigure[ ]{
          \label{fig:energy_f}
          \includegraphics[width=0.45\linewidth]{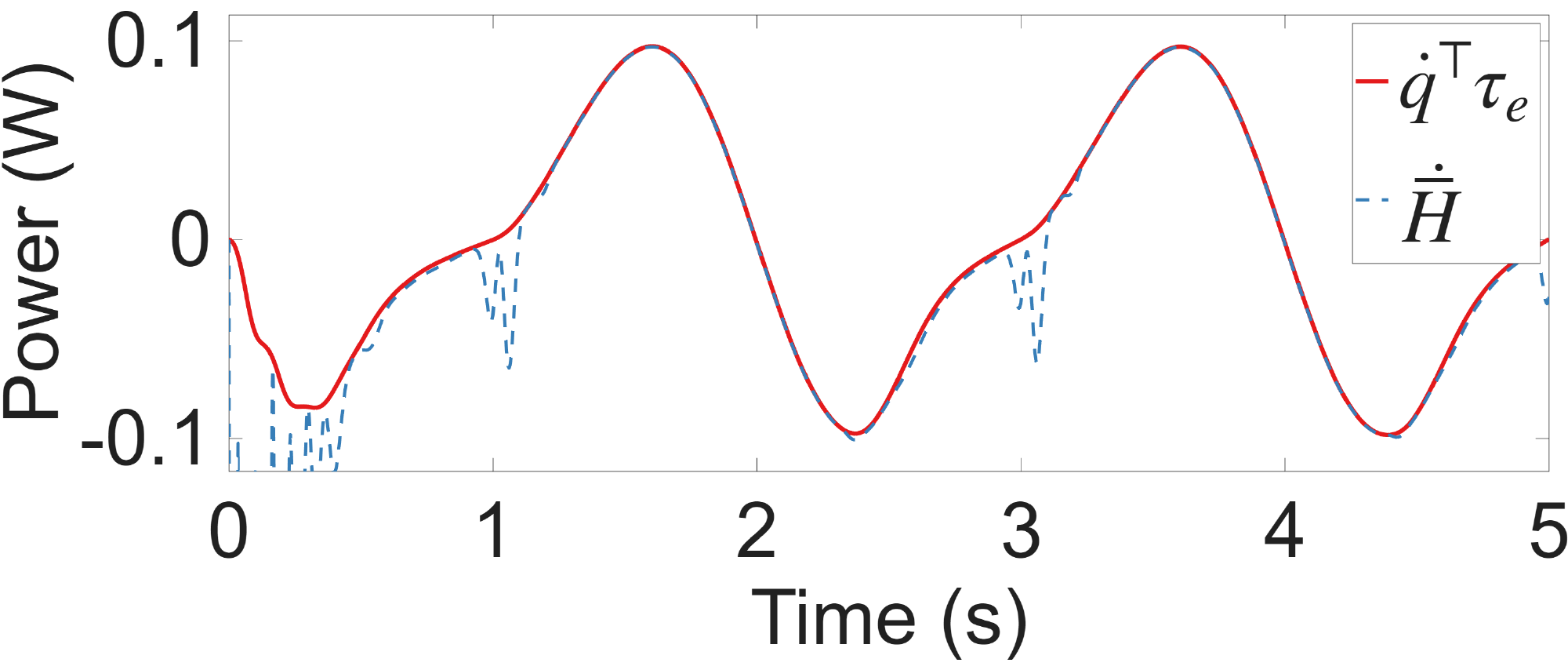}
          } 
      \subfigure[ ]{
          \label{fig:tracking_o}
          \includegraphics[width=0.45\linewidth]{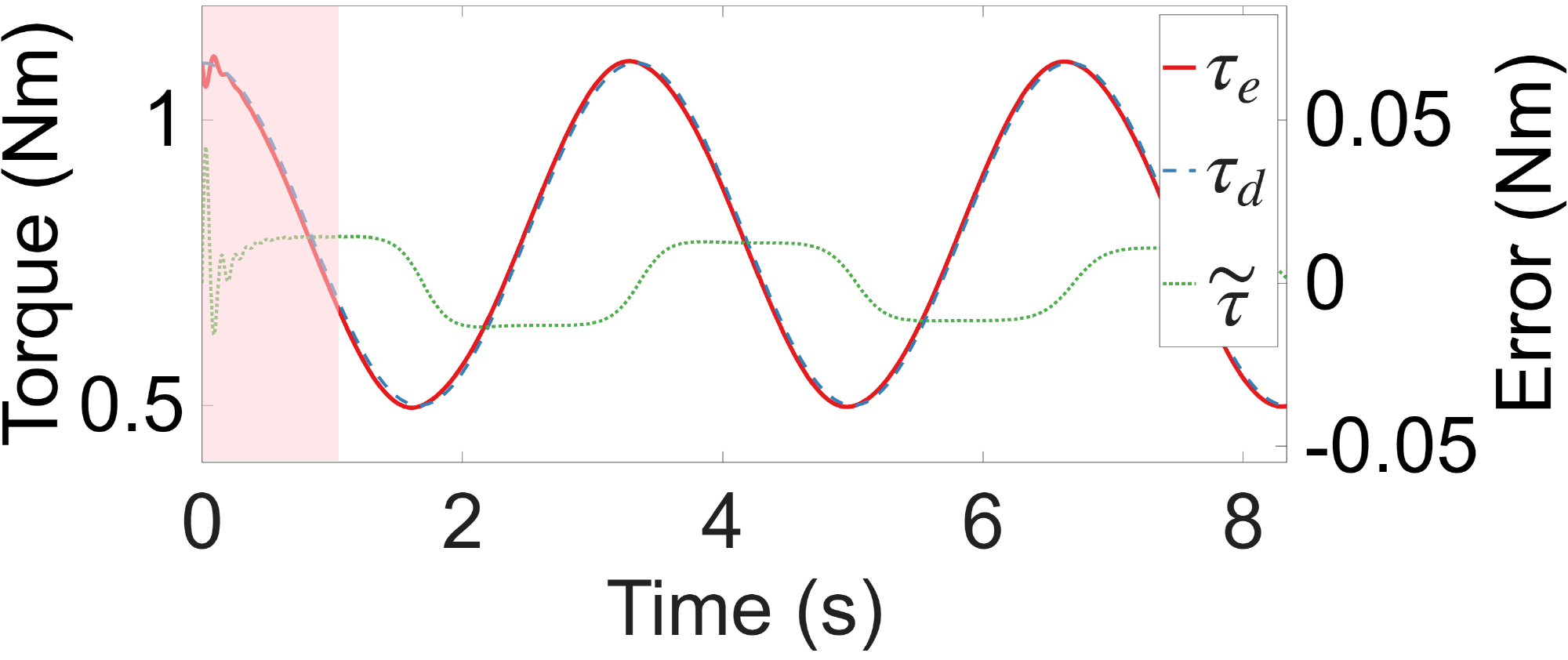}
          }\hfill
      \subfigure[ ]{
          \label{fig:energy_o}
          \includegraphics[width=0.45\linewidth]{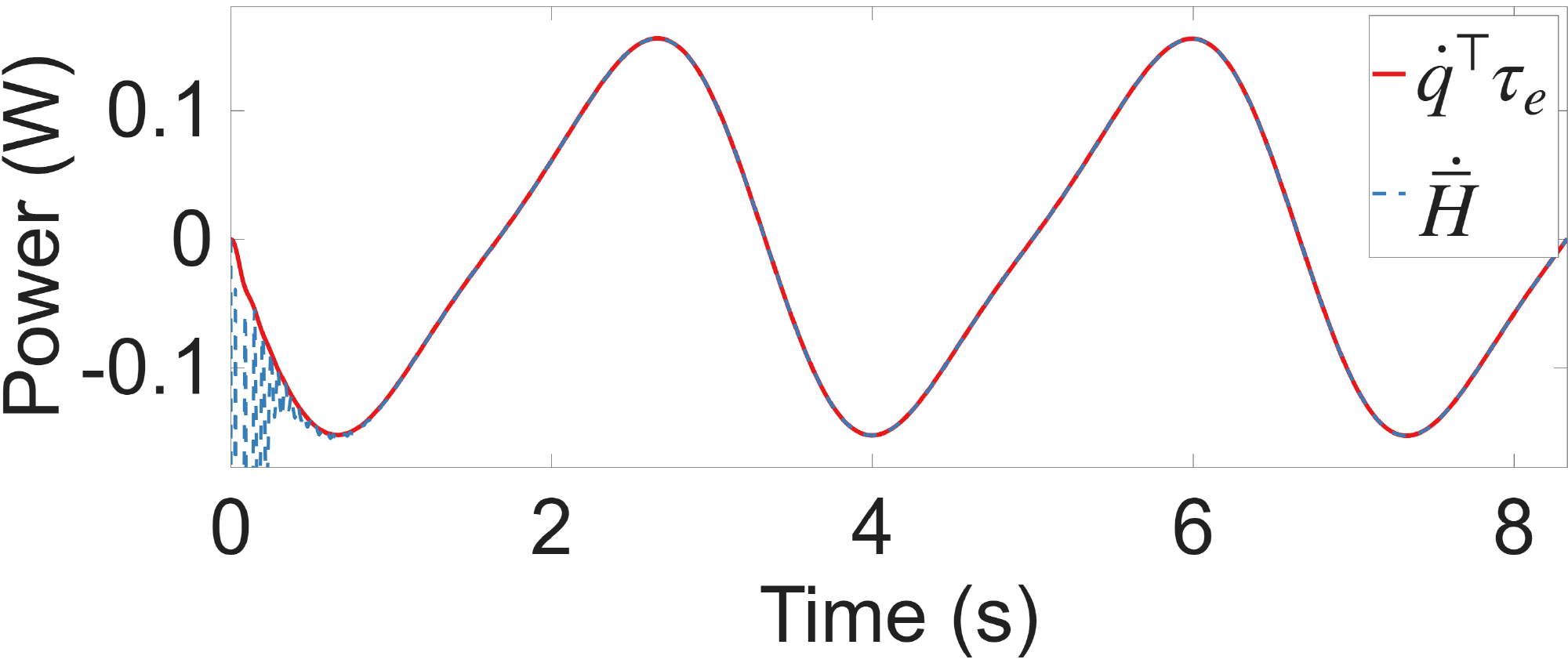}
          } 
\caption{\textbf{Simulation results of tracking performance and system power variations when depleted tanks are activated}, represented by shaded areas: (a)–(b) torque tank activation; (c)–(d) observer tank activation.}
\label{fig:tank_track}
\vspace{-0.15cm}
\end{figure}

\begin{figure}[!t]
    \centering
    \includegraphics[width=0.95\linewidth]{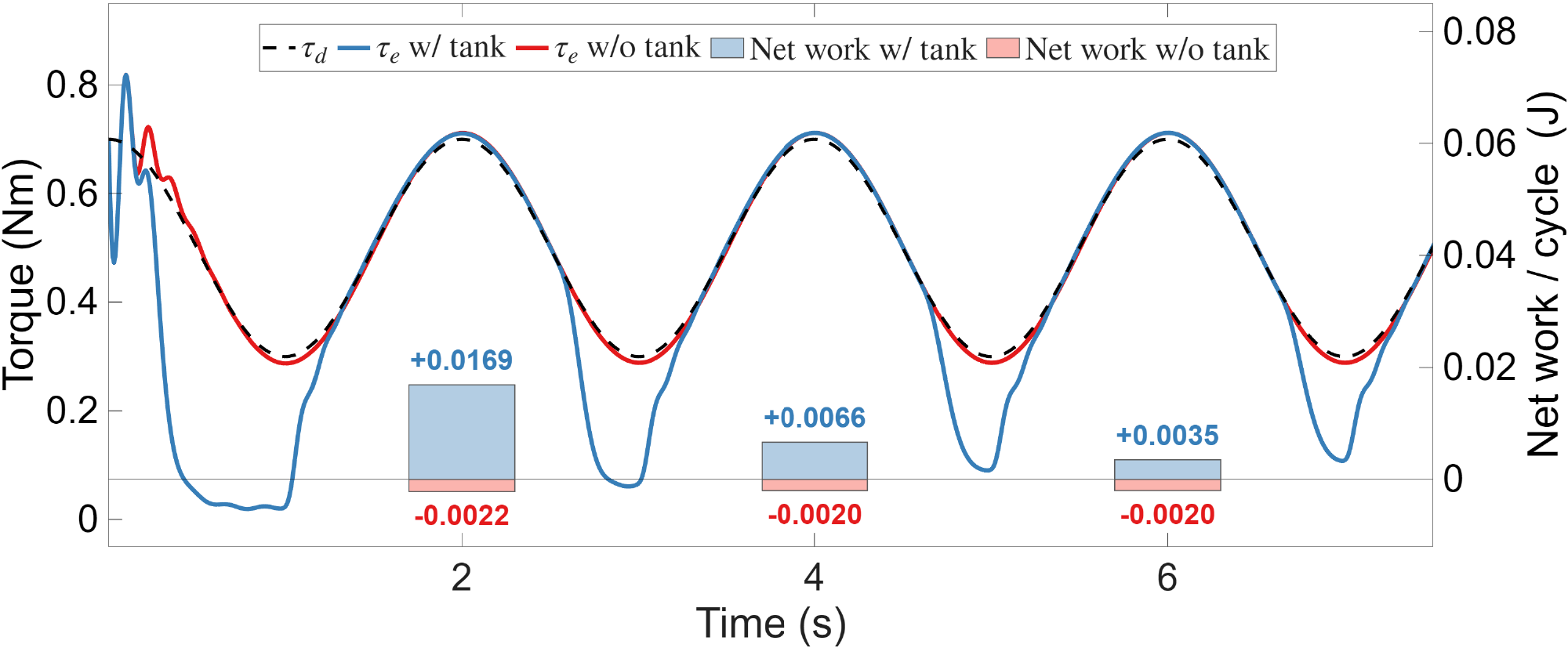}
    \caption{\textbf{Ablation study on the energy tank.} Torque-tracking performance is compared with and without the energy tank under periodic torque commands. The net work injected into the joint in each cycle is also evaluated, where a positive value indicates that the system absorbs external energy, ensuring passivity.}
    \label{fig:simu_ablation}
\end{figure}

For the simulation, the torque command parameters are set to $\bm K_p=0.5$ and $\bm K_d=0.05$. The passivity-based controller is parameterized with $\bm K_v=5$, $\bm K_s=2.0$, $\bm K_{\theta}=1.0$, $\alpha_{\theta}=5.0$, $\bm K_e=4.0$, and $\beta_e=30$, together with an SEA stiffness of $\bm K=2.0~\mathrm{Nm/rad}$. The selected parameters satisfy conditions~(\ref{equ:con1})--(\ref{equ:con3}).
When the energy tanks are not activated, the proposed controller achieves accurate torque tracking, as shown in Fig.~\ref{fig:good_force_tracking}, with the tracking error converging to zero. When the tanks are activated under conditions where the stored energy is insufficient to compensate for external energy injection, the controller sacrifices torque-tracking performance to preserve passivity. In Fig.~\ref{fig:tank_track}, the shaded areas indicate the separate activation of the torque tank and the observer tank. When the activated tank is depleted, the tracking performance is intentionally reduced to avoid passivity violation. Conversely, when passivity can be maintained without tank activation, the controller prioritizes the torque-tracking task. Throughout the process, including controller mode transitions, the passivity of the system is guaranteed, as demonstrated in Fig.~\ref{fig:tank_track}.
An ablation study on the energy tank is conducted by comparing periodic torque-tracking performance with and without the energy tank. As shown in Fig.~\ref{fig:simu_ablation}, the values above the bars represent the net work injected into the system during each cycle. When the energy tank is enabled, the controller relaxes torque-tracking performance to preserve passivity. In contrast, without the energy tank, the controller releases energy into the environment in each cycle, which may compromise the safety of human--robot interaction.

\begin{figure}[!t]
    \centering
    \includegraphics[width=0.95\linewidth]{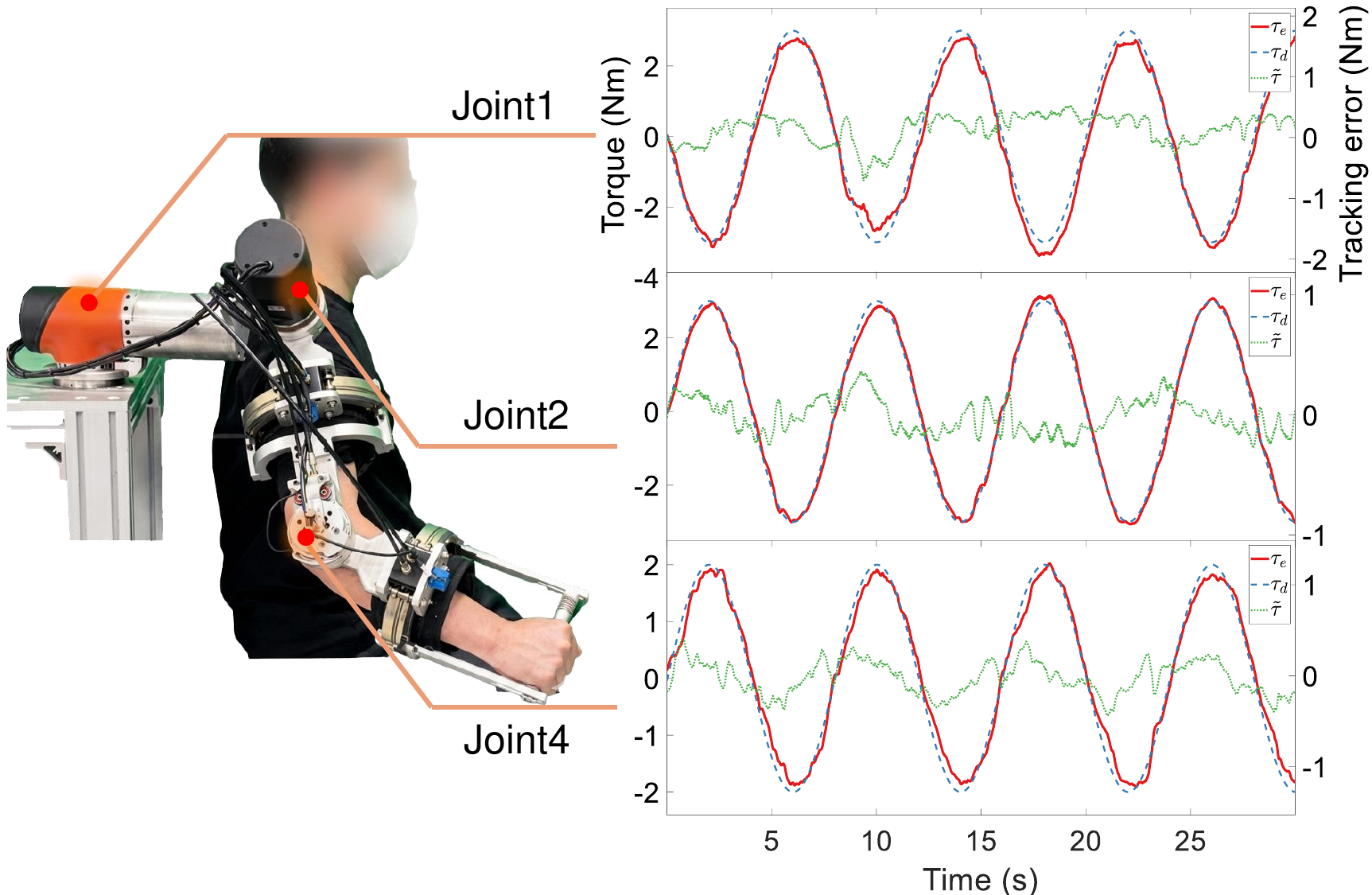}
    \caption{\textbf{Interaction torque tracking achieved by implementing the proposed controller on the exoskeleton.} The controller is applied to two direct-drive shoulder joints, Joints 1 and 2, and one cable-driven compliant joint, Joint 4.}
    \label{fig:real_tracking}
    \vspace{-0.3cm}
\end{figure}

\begin{figure}[!t]
    \centering
    \includegraphics[width=0.95\linewidth]{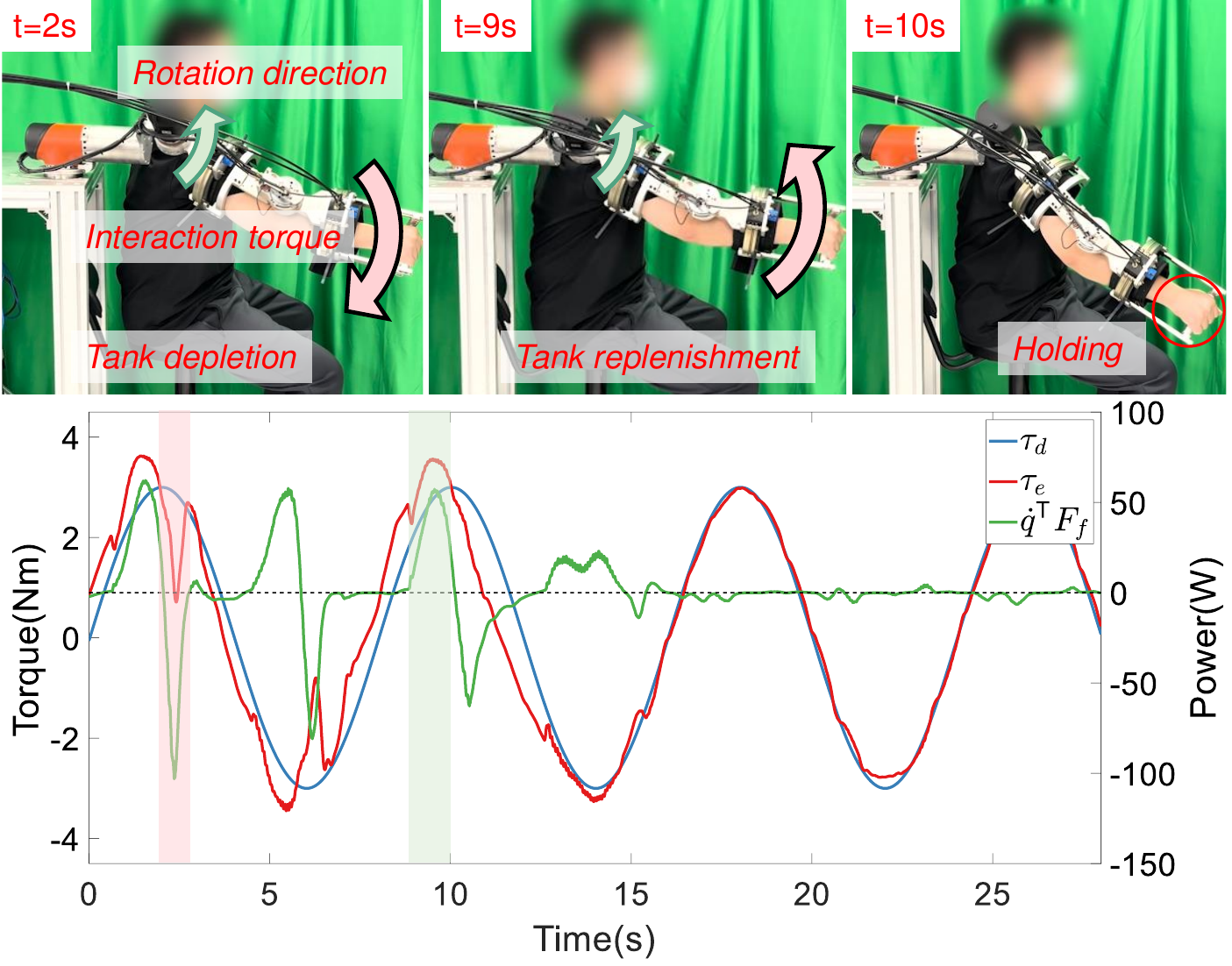}
    \caption{\textbf{Torque-tracking performance of the shoulder joint under tank state transition switching.} The red shaded region represents tank depletion, while the green shaded region indicates that the wearer interacts with the exoskeleton to replenish tank, allowing the controller to switch back to torque tracking as the primary objective.}
    \label{fig:passiv_shoulder}
\end{figure}


We implemented the proposed controller on the custom-built exoskeleton. The joints responsible for internal and external rotation, namely Joints 3 and 5, are loosely coupled to the wearer and may lose contact during motion, which would prevent stable human--robot interaction. Therefore, assistance was provided only by the direct-drive Joints 1 and 2 for shoulder motion and cable-driven compliant Joint 4 for elbow motion.
The torque-tracking performance of these activated joints is shown in Fig.~\ref{fig:real_tracking}. The tracking RMSEs are 0.27 $\mathrm{Nm}$, 0.14 $\mathrm{Nm}$, and 0.18 $\mathrm{Nm}$ for Joints 1, 2, and 4, respectively.

\begin{figure}[!t]
    \centering
    \includegraphics[width=0.95\linewidth]{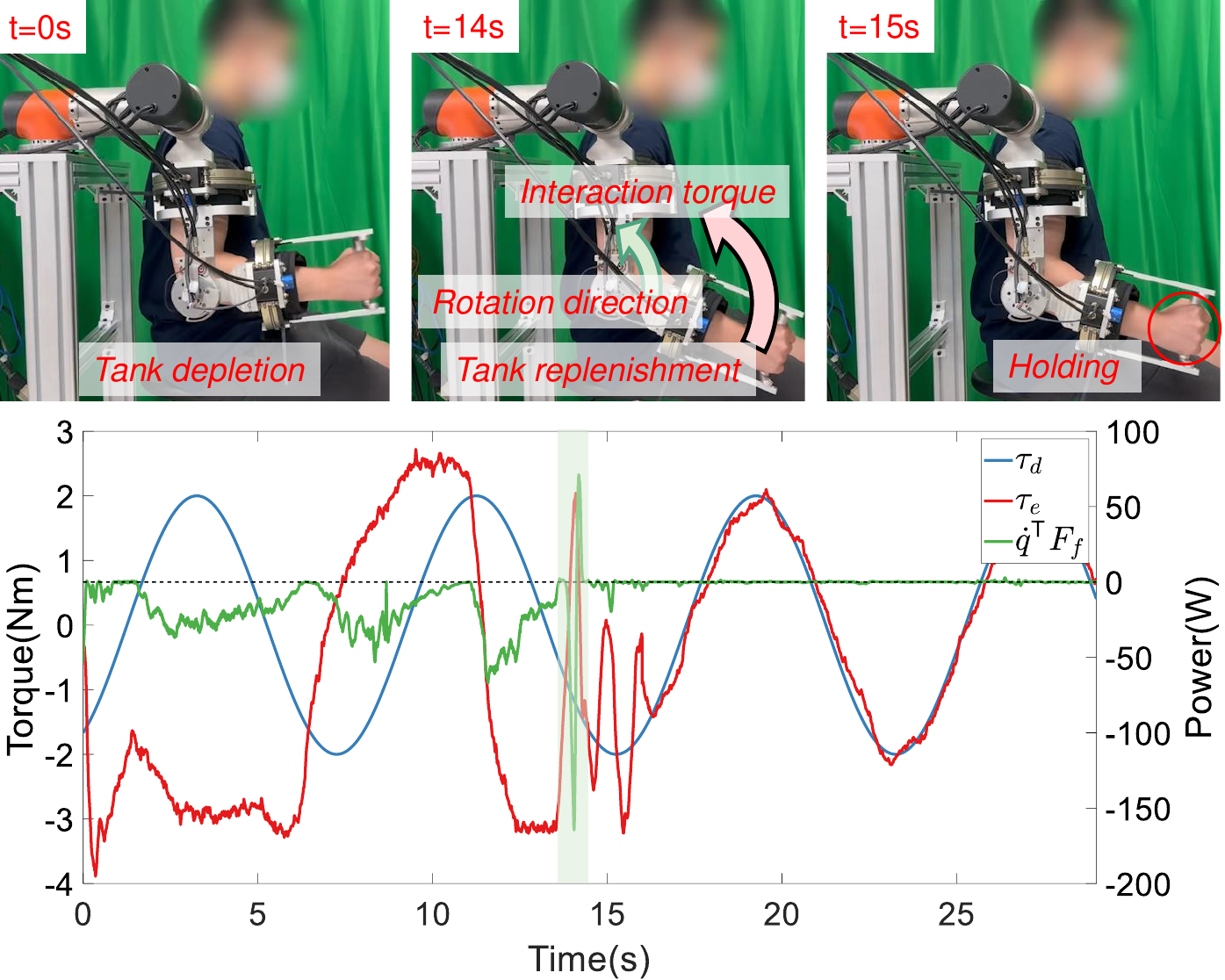}
    \caption{\textbf{Torque-tracking performance of the cable-driven elbow joint under tank state transition switching.} The green shaded region indicates the phase in which the wearer interacts with the exoskeleton to replenish tank, allowing the controller to switch back to torque tracking as the primary objective.}
    \label{fig:passiv_elbow}
    \vspace{-0.55cm}
\end{figure}

We further evaluated the controller performance during transitions between different energy-tank states. 
The controller-switching scenarios were demonstrated on two representative joints: Joint 2, corresponding to the direct-drive shoulder joints, and Joint 4, corresponding to the cable-driven compliant elbow joint. During each test, the remaining joints were deactivated, allowing the wearer to focus on interacting with the joint and thus trigger the controller priority switch.
The snapshots and interaction torque-tracking performance for Joint 2 under passivity-based controller switching are shown in Fig.~\ref{fig:passiv_shoulder}. 
The red shaded region indicates energy-tank depletion. During this phase, the wearer drives the exoskeleton such that the interaction torque acts opposite to the direction of joint rotation, thereby consuming the energy stored in the tank.
Once the tank is depleted, the controller relaxes the torque-tracking objective to preserve passivity. 
The green shaded region represents the subsequent phase in which the energy tank is replenished and the wearer then holds the exoskeleton to reduce tank energy dissipation, thereby allowing the controller to return to its intended task of torque tracking.
For Joint~4, a cable-driven elbow joint, the energy tank was initialized at a low energy level to represent a depleted state. The corresponding interaction results are shown in Fig.~\ref{fig:passiv_elbow}.
The exoskeleton initially provided only compensation for system dynamics and cable friction, allowing it to follow the wearer’s motion while preserving passivity. At approximately 14 s, the wearer interacted with the joint such that the interaction torque was aligned with the joint rotation, as indicated by the green shaded region. This interaction indirectly replenished the energy tank and triggered the controller to prioritize interaction torque tracking. After the controller behavior switched, the wearer actively held the handle to reduce tank energy dissipation, thereby maintaining torque tracking as the primary control objective.
By implementing the proposed passivity-based controller, the exoskeleton can provide refined upper-limb movement assistance while ensuring the safety of human--robot interaction.

\subsection{Anatomical Assistance}
We implemented the proposed ATP controller on our upper-limb exoskeleton. To generate the future reference torque sequence required for the subsequent desired-torque optimization in (\ref{opt_all}), we employed the intention predictor and anomaly detector developed in~\cite{chen2025upper}. The motion intentions of Joints~1, 2, and 4 were predicted over a $70~\mathrm{ms}$ horizon, while the wearer’s interaction comfort was assessed using a diffusion-based anomaly score, denoted by $s$.
To compensate for the computational latency introduced by the inference pipeline, the torque corresponding to the $10~\mathrm{ms}$ look-ahead point in the optimized sequence was applied to the exoskeleton. This look-ahead interval was empirically selected to ensure that the assistance remained both comfortable for the wearer and sufficiently responsive.

\begin{figure}[!t]
    \centering
    \includegraphics[width=0.9\linewidth]{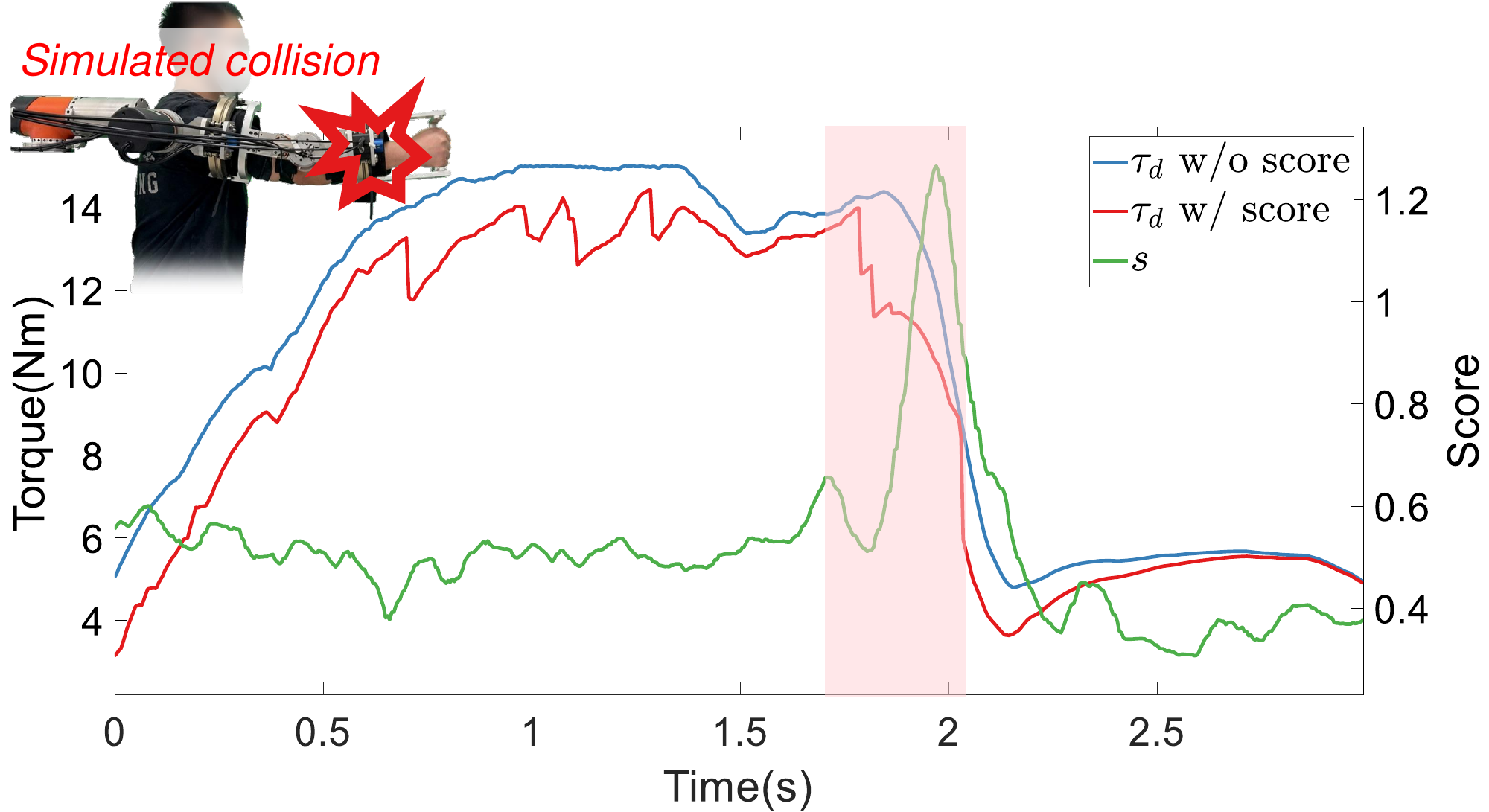}
    \caption{\textbf{Torque refinement during the simulated collision.} The red shaded region indicates the period during which the collision occurs, while the refined torque decreased accordingly to mitigate conflict.}
    \label{fig:anom_coli}
\end{figure}

\begin{figure}[!t]
    \centering
    \includegraphics[width=0.9\linewidth]{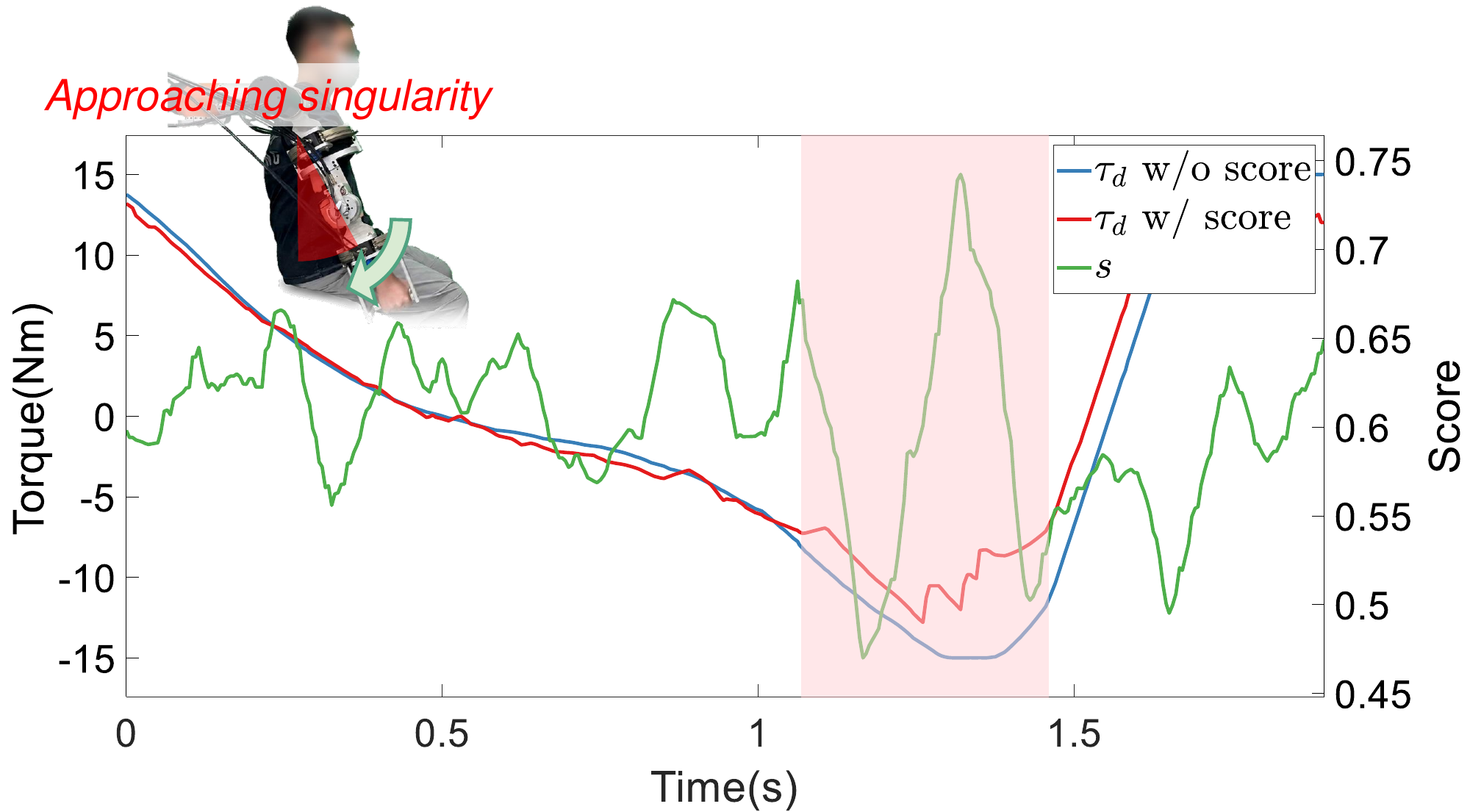}
    \caption{\textbf{Torque refinement during motion toward a kinematic singularity.} The red sector denotes the region approaching Jacobian singularity, while the red shaded area indicates the period during which the anomaly occurs. The refined torque increases accordingly, generating a corrective bias that guides the motion away from the singularity.}
    \label{fig:anom_sing}
\end{figure}

\begin{figure*}[!t]
    \centering
    \includegraphics[width=0.95\linewidth]{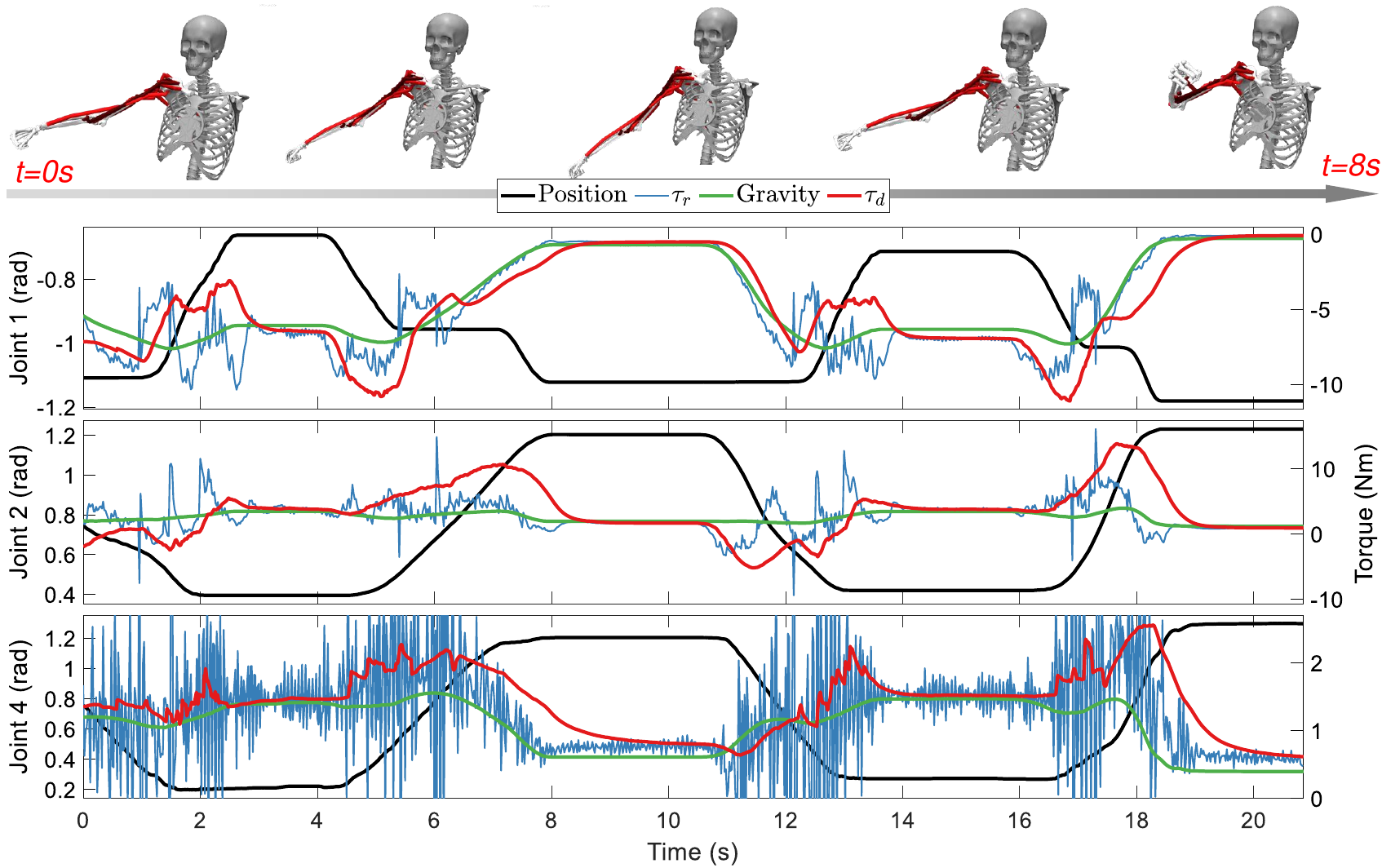}
    \caption{\textbf{Performance of torque refinement compared with different torque profiles.} The wearer performs a compound movement involving three joints, with activated tendons highlighted in dark red. Torque refinement improves the smoothness and safety of the delivered assistance while distinguishing between static and dynamic limb states and responding promptly to provide movement-dependent anatomical assistance.}
    \label{fig:refinement}
\end{figure*}

Previous work has demonstrated the feasibility of using an anomaly score to detect abnormal conditions during various upper-limb movements. In this study, we incorporated the anomaly score into the online torque-refinement framework and extended its application to two representative scenarios: a simulated collision and motion near a kinematic singularity.
In the simulated-collision experiment, the wearer was instructed to rapidly reverse the direction of arm motion to emulate an unexpected impact with an obstacle. Figure~\ref{fig:anom_coli} compares the refined torques obtained with and without anomaly-score guidance. The red shaded region indicates the period during which the simulated collision occurred. As the anomalous interaction emerged, the anomaly score increased accordingly. Meanwhile, the score-guided refinement substantially reduced the assistance torque compared with the result obtained without anomaly-score guidance, thereby mitigating the human--robot interaction conflict.
Since the limb pose is measured in the exoskeleton coordinate whereas the torque reference is generated in the musculoskeletal coordinate, the Jacobian used to map between the two systems is ill-conditioned. In the present setup, a singularity occurs when Joint~1 and Joint~2 simultaneously approach $0$. The corresponding results are shown in Fig.~\ref{fig:anom_sing}. The red sector denotes the singular region in the joint configuration space, while the red shaded region indicates the period during which the wearer’s arm entered this region. As the arm approached the singularity, the score-guided refinement reshaped the assistance torque to generate a corrective bias that directed the motion away from the singular region.
By incorporating the anomaly score into online torque refinement, the proposed method can identify different anomalous interaction conditions and adapt the assistance torque accordingly, thereby improving wearer comfort and interaction safety.

\begin{figure*}[!t]
    \centering
    \includegraphics[width=0.95\linewidth]{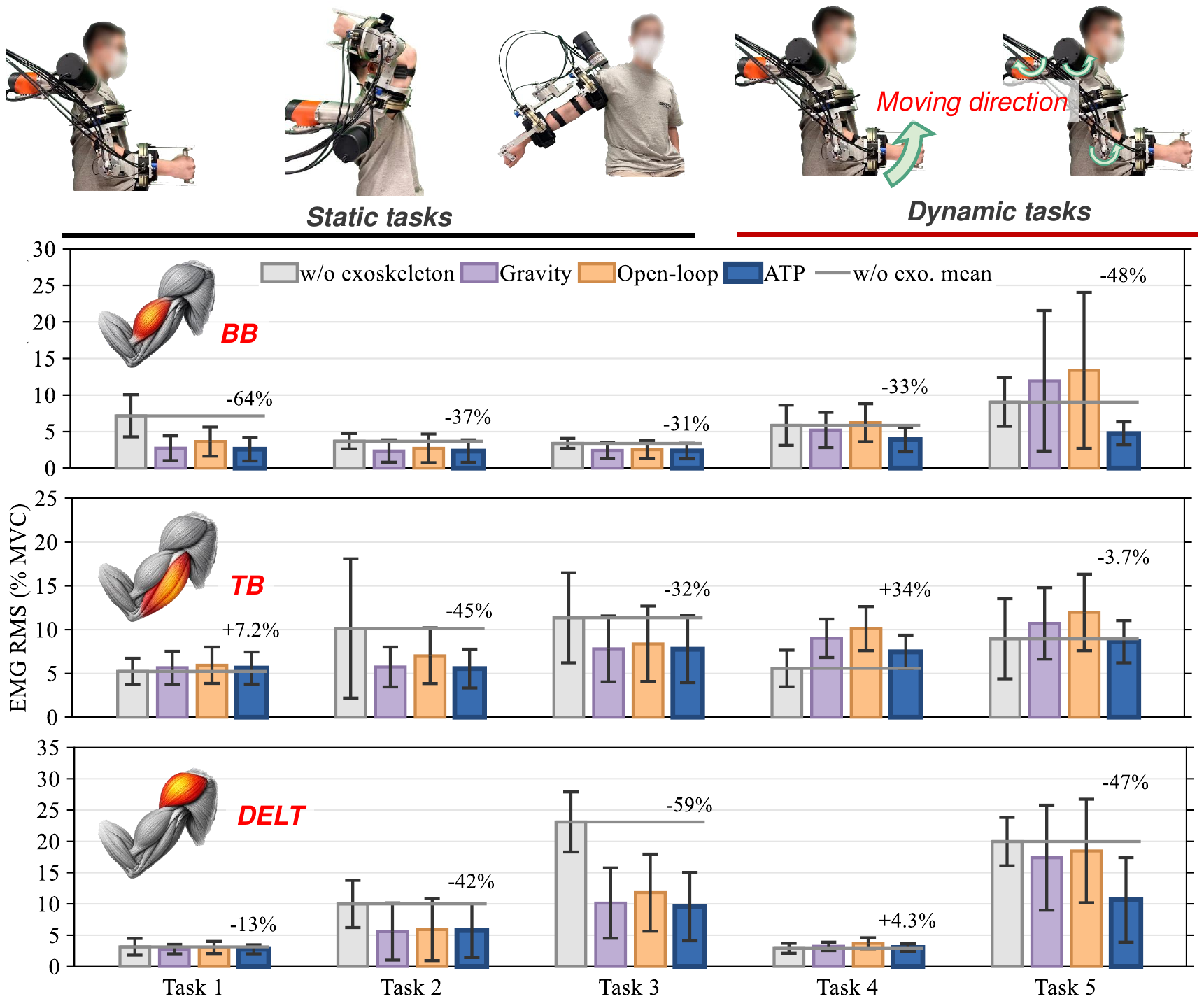}
    \caption{\textbf{EMG responses of different muscles under various assistance conditions across multiple tasks.} The proposed ATP controller assists the wearer during three static tasks and two dynamic tasks, reducing the EMG activity of the primary muscles engaged in each task.}
    \label{fig:EMG}
\end{figure*}

To validate whether the online torque refinement module captures the dynamic characteristics of limb motion and responds proactively to provide anatomical assistance, we conducted an ablation study comparing three torque signals: the raw reference torque $\bm{\tau}_r$, the gravity-compensation torque computed from the instantaneous limb configuration, and the desired torque $\bm{\tau}_d$ generated by the proposed ATP controller. 
The actual dynamic joint torque required to support limb motion was not included as a ground-truth baseline because the joint accelerations required for inverse dynamics are highly susceptible to noise and cannot be estimated both accurately and rapidly with the current experimental setup.
A compound movement involving three joints was selected as the testing protocol, as shown in Fig.~\ref{fig:refinement}. The first row presents a replay of the first $8~\mathrm{s}$ of the complete movement in the musculoskeletal simulation, where the activated tendons are highlighted in dark red. The corresponding torque profiles are also shown in Fig.~\ref{fig:refinement}. 
Among the three signals, the raw reference torque exhibits the most pronounced spikes, arising from nonlinear tendon dynamics and the inherent variability of the neural network-based muscle controller. 
Although this torque can drive the musculoskeletal system to accurately track the prescribed joint trajectories in simulation, its excessive oscillations make it unsuitable for direct application as an assistance command to the wearer.
During static phases, both the raw reference torque and the refined desired torque converged toward the gravity-compensation torque, indicating that the ATP controller can identify static conditions and generate joint torques consistent with the expected gravitational load. During dynamic motion, the refined torque increased or decreased according to the movement requirements. For example, it increased to assist arm elevation and adapted to support the limb weight during arm adduction. These results demonstrate that the optimized torque can respond promptly to changes in limb dynamics and provide movement-dependent anatomical assistance.

\begin{table}[t!]
\caption{Characteristics of Subjects}
\centering
\begin{tabular}{cccccc} 
\toprule
Subject & Gender & Age (y) & Weight (kg) & Height (cm)  \\ 
\midrule
1 &  Male   & 26   & 67        & 172       \\
2   & Male   & 24   & 70        & 170      \\
3   & Male   & 22   & 71        & 174      \\
4   & Male   & 37   & 73        & 167      \\
5   & Female   & 20   & 58        & 169      \\
\bottomrule
\end{tabular}
\label{tab:subpttri}
\end{table}

We conducted a comparative study involving four experimental conditions: unassisted movement without wearing the exoskeleton, gravity-compensation assistance derived from the musculoskeletal model, open-loop assistance without the interaction torque controller, and assistance provided by the proposed ATP controller. The performance under each condition was evaluated using EMG signals recorded from the BB, TB, and DELT muscles in five recruited participants. The participant characteristics are summarized in Table~\ref{tab:subpttri}. Before participating in the experiments, all participants were fully informed of the experimental procedures and provided written informed consent.
For each participant, the magnitude of anatomical assistance was scaled according to body weight. A participant weighing $70~\mathrm{kg}$ was assigned a baseline assistance level of $30\%$ of $\bm{\tau}_d$, with the assistance for other participants adjusted proportionally. Before muscle activity was measured during the experimental tasks, each participant performed a maximum voluntary contraction test for each monitored muscle.
Five upper-limb tasks were included, comprising three static tasks and two dynamic tasks, as illustrated in Fig.~\ref{fig:EMG}. The static tasks consisted of holding the forearm horizontally to engage the BB, holding the arm in an elevated posture to engage the TB, and maintaining the arm in a laterally raised posture to engage the DELT. The dynamic tasks comprised elbow flexion and a coordinated three-joint movement. Each participant performed three trials for every combination of task and assistance condition, and the resulting data were statistically analyzed across all trials.

The experimental results are summarized in Fig.~\ref{fig:EMG}. Across all static tasks, ATP assistance and gravity-compensation assistance produced similar EMG responses because both methods converged to approximately the same assistance torque under static conditions. By supporting the weight of the limb, the proposed ATP controller substantially reduced the EMG activity of the primary muscles involved in each task. Moreover, because anatomical assistance was applied across all assisted joints, EMG activity in the other monitored muscles either remained comparable to or decreased from that observed without the exoskeleton. Specifically, relative to the condition without the exoskeleton, ATP assistance reduced BB activity by $64\%$ in Task~1, TB activity by $45\%$ in Task~2, and DELT activity by $59\%$ in Task~3.
For the dynamic tasks, the anatomical assistance generated by ATP captured the dynamic characteristics of the musculoskeletal model and responded promptly during limb motion. Consequently, the EMG activity of the muscles involved in lifting the limb was reduced relative to the condition without the exoskeleton. In contrast, gravity compensation and open-loop assistance could not adapt promptly to changes in limb motion and therefore occasionally impeded the movement, in some cases resulting in increased EMG activity. 
Note that during the arm-lowering phase, the cable-driven elbow joint exhibited a relatively low rotational speed because of its structural limitations. Consequently, participants tended to push against the handle instinctively, resulting in increased TB activity. 
Despite this limitation, the proposed ATP controller reduced BB activity by $33\%$ in Task~4 and reduced BB and DELT activity by $48\%$ and $47\%$, respectively, in Task~5.
Overall, the proposed ATP controller provided effective anatomical assistance across different tasks and reduced the EMG activity of the primary muscles compared with the other assistance conditions.


\section{Discussion and Conclusions}
\subsection{Limitations}
Despite the demonstrated effectiveness of the proposed framework, several limitations remain and point to directions for future work.
\begin{enumerate}
\item[1)] The anatomical reference torque is produced by a single generic MyoArm model driven solely by the wearer's measured kinematics. As a result, the current framework targets unloaded, free-space movements and does not represent muscle behavior under external hand loads. Incorporating end-effector load sensing and load-dependent muscle coordination could extend the method to payload-handling tasks.
\item[2)] Because anatomical assistance is delivered through the passivity-based controller, the torque-tracking objective is relaxed once the energy tank is depleted in order to preserve system passivity, and restoration of the desired assistance then relies on the wearer's interaction to replenish the tank. Investigating mechanisms that guarantee a minimum level of assistance even under tank depletion is a worthwhile direction.
\item[3)] The method was evaluated on a small cohort of five able-bodied participants, and assistance was restricted to three joints owing to the mechanical constraints of the current prototype. Evaluation on a larger and more diverse population—including intended target users—together with an exoskeleton providing additional actuated degrees of freedom would enable a more comprehensive assessment of the method's effectiveness and generalizability.
\end{enumerate}

\subsection{Conclusions}
\noindent\textbf{Intellectual Merits}: 
This paper presented ATP, a framework for delivering safe anatomical assistance with upper-limb exoskeletons by integrating a learning-based biomechanical reference with passivity-based interaction control. Unlike existing anatomical assistance approaches, which have largely focused on periodic weight-support tasks for lower-limb exoskeletons, ATP generates anatomically informed torque references for complex, non-periodic upper-limb movements directly from a musculoskeletal model. This eliminates the need for costly biomechanical computations.
To support large-scale learning, we developed and open-sourced a GPU-accelerated muscle–tendon simulation environment and trained a unified muscle controller to generate biomechanically informed assistance across diverse upper-limb movements. An online torque-refinement scheme subsequently adapts the generated reference to the wearer’s motion and interaction state, while a passivity-based controller ensures safe and compliant torque delivery.
By combining biomechanical learning with control-theoretic safety in a unified framework, ATP enables movement-dependent anatomical assistance for upper-limb exoskeletons. Its effectiveness was demonstrated through simulations and real-world experiments, in which the exoskeleton delivered responsive assistance and reduced the activity of the primary muscles during both static and dynamic upper-limb tasks.

\noindent\textbf{Potential Impacts}: Because ATP infers assistance from proprioceptive signals and wearable inertial sensing rather than costly biomechanical instrumentation, it lowers the barrier to deploying anatomically informed assistance outside laboratory settings. This capability has potential applications in industrial support, rehabilitation, and assistance with activities of daily living. The passivity guarantee, established at the human--robot interaction port and maintained through energy-tank regulation, also provides a principled basis for safe physical human--robot interaction that may extend to other compliant wearable robotic systems.
Moreover, the open-source musculoskeletal simulation environment provides the research community with a practical platform for large-scale, high-fidelity muscle–tendon learning and may accelerate the development of embodied musculoskeletal control methods beyond exoskeleton applications.
By demonstrating that detailed, movement-dependent anatomical assistance can be generated efficiently and delivered safely, this work advances the development of generalized assist-as-needed upper-limb exoskeletons capable of adapting to a wide range of wearers, tasks, and real-world conditions.





{\small
\bibliographystyle{ref/IEEEtran}
\bibliography{ref/IEEEabrv, ref/ref}
}

\end{document}